\documentclass{article}
\usepackage{graphicx}
\usepackage{subcaption}
\usepackage{amsmath}
\usepackage{tikz}
\usepackage{pgfplots}
\usetikzlibrary{spy, backgrounds}

\usepackage[preprint]{neurips_2023}

\usepackage[utf8]{inputenc} 
\usepackage[T1]{fontenc}    
\usepackage{hyperref}       
\usepackage{url}            
\usepackage{booktabs}       
\usepackage{amsfonts}       
\usepackage{nicefrac}       
\usepackage{microtype}      
\usepackage{xcolor}         

\title{Probing In-Context Learning: Impact of Task Complexity and Model Architecture on Generalization and Efficiency}

\author{
  Binwen Liu\thanks{Corresponding author: binwenliu.ai@gmail.com } \\
  University of California, Berkeley \\
  \texttt{liubinwen@berkeley.edu}
  \And
  Peiyu Xu\thanks{These authors contributed equally as second authors.\newline
  Our code and models are available at: \href{https://github.com/Binwen6/CS182_PROJECT_2025}{\url{https://github.com/Binwen6/CS182_PROJECT_2025}}} \\
  University of California, Berkeley \\
  \texttt{xupeiyu@berkeley.edu} \\
  \And
  Quan Yuan\footnotemark[2] \\
  University of California, Berkeley \\
  \texttt{yquan@berkeley.edu} \\
  \And
  Yihong Chen\footnotemark[2] \\
  University of California, Berkeley \\
  \texttt{cyh1102@berkeley.edu} \\
}

\begin{document}

\maketitle

\begin{abstract}
We investigate in-context learning (ICL) through a meticulous experimental framework that systematically varies task complexity and model architecture. Extending beyond the linear regression baseline, we introduce Gaussian kernel regression and nonlinear dynamical system tasks, which emphasize temporal and recursive reasoning. We evaluate four distinct models: a GPT2-style Transformer, a Transformer with FlashAttention mechanism, a convolutional Hyena-based model, and the Mamba state-space model. Each model is trained from scratch on synthetic datasets and assessed for generalization during testing. Our findings highlight that model architecture significantly shapes ICL performance. The standard Transformer demonstrates robust performance across diverse tasks, while Mamba excels in temporally structured dynamics. Hyena effectively captures long-range dependencies but shows higher variance early in training, and FlashAttention offers computational efficiency but is more sensitive in low-data regimes. Further analysis uncovers locality-induced shortcuts in Gaussian kernel tasks, enhanced nonlinear separability through input range scaling, and the critical role of curriculum learning in mastering high-dimensional tasks.
\end{abstract}

\section{Introduction}

In-context learning (ICL) has emerged as a powerful paradigm in machine learning, enabling models to adapt to new tasks with minimal supervision by leveraging contextual information. Recent studies have framed ICL through the lens of meta-learning, where models learn to approximate functions from a distribution over tasks using only contextual supervision~\citep{dong2022survey}. While foundational work has demonstrated the ability of transformers to internalize simple learning algorithms for tasks like linear regression~\citep{garg2023transformerslearnincontextcase}, the scope of these investigations has often been limited to specific architectures and function classes.

This project extends the study of ICL along two critical dimensions: function complexity and model generality. First, we incorporate more complex function families, such as Gaussian kernel regression and nonlinear dynamical systems, which introduce challenges related to smoothness, locality, and temporal dependencies. These function classes push the boundaries of ICL beyond simpler tasks previously explored. Second, we evaluate ICL performance across a diverse set of models: a baseline GPT2-style transformer, a transformer variant with FlashAttention~\citep{dao2022flashattention}, a Hyena-based attention-free model~\citep{poli2023hyena}, and Mamba, a state-space model with selective recurrence mechanisms~\citep{gu2023mamba}. 

By exploring this expanded landscape, we aim to uncover how architectural choices influence generalization in ICL settings. Our findings will provide insights into the strengths and limitations of different architectures when confronted with increasingly complex learning tasks, ultimately guiding the development of more robust and versatile ICL systems.

\section{Related Work}

The study of in-context learning (ICL) has been significantly shaped by the meta-learning perspective, which views ICL as a process where models learn to approximate functions from a task distribution using contextual supervision. A comprehensive survey by Dong et al. (2022)~\citep{dong2022survey} outlines the definitions, techniques, and applications of ICL, emphasizing its role in enabling few-shot learning without parameter updates.

Foundational work by Garg et al. (2023)~\citep{garg2023transformerslearnincontextcase} established a framework for evaluating ICL using synthetic function families, such as linear regression and Fourier approximation. Their results showed that transformers can effectively internalize simple learning algorithms, but their analysis was constrained to a narrow set of architectures and function classes. Subsequent studies have expanded the scope of ICL to more diverse and complex function families. For instance, Cole et al. (2025)~\citep{cole2025icllds} explored ICL in linear dynamical systems, while Bhattamishra et al. (2023)~\citep{bhattamishra2023} investigated its applicability to Boolean functions. Additionally, Sun et al. (2025)~\citep{sun2025polykernelicl} and Cole et al. (2024)~\citep{cole2024ellipticpdeicl} applied ICL to nonlinear kernels and elliptic partial differential equations, respectively, highlighting the growing versatility of ICL across domains and underscoring the need to understand how different model architectures perform under these conditions.

Concurrently, the development of architectures capable of handling long sequences more efficiently than traditional transformers has gained traction. FlashAttention, introduced by Dao et al. (2022)~\citep{dao2022flashattention}, addresses the computational bottlenecks of standard attention mechanisms by implementing an IO-aware exact attention algorithm, reducing memory usage and speeding up computations. The Hyena model, proposed by Poli et al. (2023)~\citep{poli2023hyena}, offers an alternative by replacing attention with subquadratic-time convolutional operations, providing improved efficiency for tasks involving long contexts. Mamba, developed by Gu et al. (2023)~\citep{gu2023mamba}, employs linear-time sequence modeling with selective state spaces, achieving state-of-the-art performance on various sequence modeling tasks, including language, audio, and genomics.

The increasing diversity of ICL applications and the emergence of novel architectures motivate our work. Earlier studies investigated ICL in decision trees, sparse linear functions, and neural networks~\citep{garg2023transformerslearnincontextcase}, while recent efforts have tackled time-dependent dynamics~\citep{cole2025icllds}, Boolean functions~\citep{bhattamishra2023}, nonlinear kernels~\citep{sun2025polykernelicl}, and partial differential equations~\citep{cole2024ellipticpdeicl}. These developments highlight the importance of evaluating how architectural inductive biases, such as recurrence in Mamba or convolution in Hyena, compare to attention-based mechanisms in complex ICL settings. Our work builds on these advancements by systematically assessing the ICL capabilities of diverse architectures across an extended range of function classes, offering a comprehensive analysis of how architecture design impacts ICL effectiveness in challenging and realistic scenarios.

\section{Approach}
\label{gen_inst}
In this section, we formalize the in-context learning (ICL) setup and describe the synthetic function families and model architectures that that we use. Our emphasis is on evaluating how various architectures internalize different function classes.
\subsection{Problem Setup}

We adopt a standard in-context learning (ICL) framework where the model is presented with a prompt $\{(x_i, y_i)\}_{i=1}^{T}$ of input-output pairs followed by a query input $x_{T+1}$. The model processes the full sequence $[(x_1, y_1), \dots, (x_T, y_T), x_{T+1}]$ as a single input and is tasked with predicting $y_{T+1}$. No parameter updates occur during inference; the model must generalize in-context from the prompt via forward computation.

This setup follows the meta-learning perspective of ICL, where the model implicitly learns a distribution over tasks and adapts to unseen functions on-the-fly, as discussed in Garg et al.~\cite{garg2023transformerslearnincontextcase}.

\subsection{Function Families}

To evaluate ICL generalization, we define a set of synthetic task families \(\mathcal{F}\), each representing a distribution over real-valued functions \(f: \mathbb{R}^d \to \mathbb{R}\). Each sampled function generates a prompt and query for training and evaluation, following established practices in ICL research \citep{garg2023transformerslearnincontextcase}.

\paragraph{Linear Regression.}
Each task samples a weight vector \(w \sim \mathcal{N}(0, I_d)\), normalized to unit norm. Inputs \(x_i \in \mathbb{R}^d\) are sampled from \(\mathcal{N}(0, I_d)\). The output is generated via:
\[
y_i = \langle w, x_i \rangle + \varepsilon_i, \quad \varepsilon_i \sim \mathcal{N}(0, \sigma^2),
\]
where \(\sigma \in \{0.1, 0.5, 1.0\}\) is a fixed noise level, chosen to test robustness across varying noise conditions. This formulation, standard in regression tasks, is inspired by statistical learning \citep{hastie2009elements} and ICL studies \citep{garg2023transformerslearnincontextcase}.

\paragraph{Gaussian Kernel Regression.}
We define a radial basis kernel regression task with \(C\) centers \(\{c_j\}_{j=1}^C\) and weights \(\beta \in \mathbb{R}^C\) per task. Centers \(c_j \in \mathbb{R}^d\) are sampled uniformly from \([-1, 1]^d\), and weights \(\beta_j \sim \mathcal{N}(0, 1)\). For each input \(x_i \sim \mathcal{N}(0, I_d)\):
\[
y_i = \sum_{j=1}^{C} \beta_j \cdot \exp\left(-\frac{\|x_i - c_j\|^2}{2h^2}\right) + \varepsilon_i,
\]
where \(h = 1\) is the bandwidth, and \(\varepsilon_i \sim \mathcal{N}(0, \sigma^2)\) with \(\sigma = 0.1\). Outputs are normalized to unit variance per batch to ensure consistency. This task, rooted in kernel methods \citep{scholkopf2002learning}, introduces smooth nonlinearities and locality-aware structure, as explored in ICL \citep{garg2023transformerslearnincontextcase}.

\paragraph{Nonlinear Dynamical Systems.}
Each task defines a recurrence rule \(x_{t+1} = F(x_t)\) and output \(y_t = \langle v, x_t \rangle + \varepsilon_t\), where \(v \sim \mathcal{N}(0, I_d)\) is normalized, and \(\varepsilon_t \sim \mathcal{N}(0, \sigma^2)\) with \(\sigma = 0.1\). Initial states \(x_0wah \sim \mathcal{N}(0, I_d)\). The nonlinear transition \(F\) includes:
\begin{itemize}
    \item \textbf{Polynomial}: 
          \[
          F(x) = Wx + W' [x^2] + b,
          \]
          where \([x^2]_i = x_i^2\), \(W, W' \in \mathbb{R}^{d \times d}\), \(b \in \mathbb{R}^d\) are sampled from \(\mathcal{N}(0, 1)\) and normalized to ensure stable dynamics. This form introduces controlled nonlinearity \citep{sussillo2013random}.
    \item \textbf{Tanh}: 
          \[
          F(x) = \tanh(Wx + b),
          \]
          with \(W \in \mathbb{R}^{d \times d}\), \(b \in \mathbb{R}^d\) sampled from \(\mathcal{N}(0, 1)\), capturing smooth nonlinear transitions \citep{sussillo2013random}.
    \item \textbf{Logistic}: A simple nonlinear recurrence relation defined as:
      \[
      x_{t+1} = r x_t (1 - x_t),
      \]
      where \(r = 3.9\) controls the system’s behavior, exhibiting period-doubling and chaos for certain values of \(r\) \citep{strogatz2015nonlinear}.

    \item \textbf{Duffing Oscillator}: A second-order system discretized as:
          \[
          x_{t+1} = x_t + \delta \dot{x}_t, \quad \dot{x}_{t+1} = \dot{x}_t + \delta (-\alpha x_t - \beta x_t^3 - \gamma \dot{x}_t + f \cos(\omega t)),
          \]
          with \(\alpha = 1\), \(\beta = 0.1\), \(\gamma = 0.1\), \(f = 0.5\), \(\omega = 1\), and \(\delta = 0.01\) \citep{strogatz2015nonlinear}.
    \item \textbf{Van der Pol Oscillator}: Discretized as:
          \[
          x_{t+1} = x_t + \delta \dot{x}_t, \quad \dot{x}_{t+1} = \dot{x}_t + \delta (\mu (1 - x_t^2) \dot{x}_t - x_t),
          \]
          with \(\mu = 2\), \(\delta = 0.01\) \citep{strogatz2015nonlinear}.
    \item \textbf{Lorenz System}: A chaotic system discretized as:
          \[
          x_{t+1} = x_t + \delta \sigma (y_t - x_t), \quad y_{t+1} = y_t + \delta (x_t (\rho - z_t) - y_t), \quad z_{t+1} = z_t + \delta (x_t y_t - \beta z_t),
          \]
          with \(\sigma = 10\), \(\rho = 28\), \(\beta = 8/3\), \(\delta = 0.01\) \citep{strogatz2015nonlinear}.
\end{itemize}
These tasks, inspired by nonlinear dynamics \citep{strogatz2015nonlinear} and ICL studies \citep{chan2022data}, require the model to track latent states across time, highlighting architectural capacity for recurrence and memory.

\subsection{Model Architectures}

\begin{figure}
    \centering
    \includegraphics[width=0.8\textwidth]{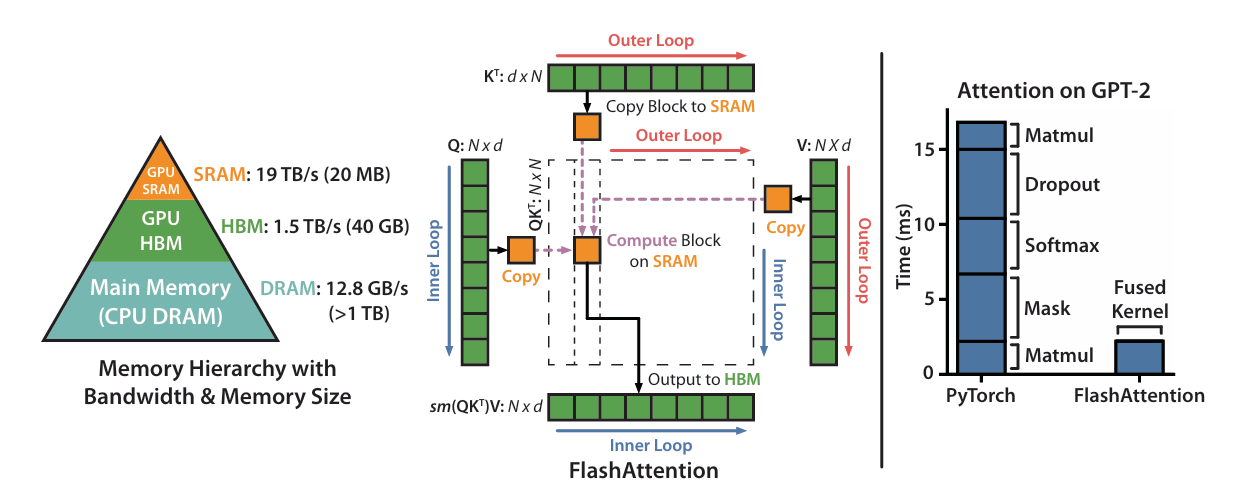}
    \caption{FlashAttention mechanism~\cite{dao2022flashattention}. The design tiles attention computation to avoid memory bottlenecks, achieving high throughput on modern hardware.}
    \label{fig:flashattention}
\end{figure}

\begin{figure}
    \centering
    \includegraphics[width=0.8\textwidth]{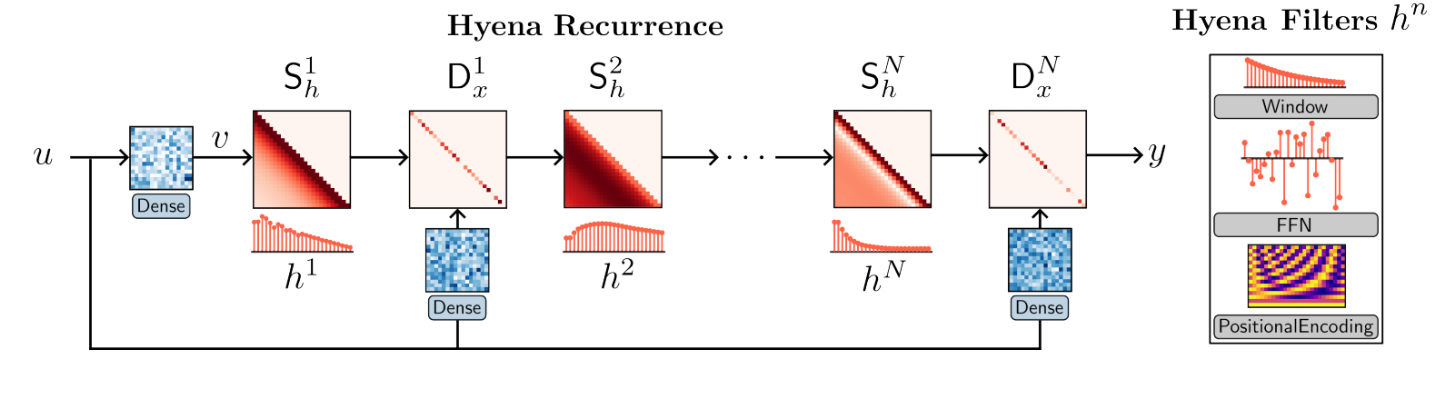}
    \caption{Hyena recurrence~\cite{poli2023hyena}. Combines implicit long convolutions with multiplicative gating, allowing attention-like behavior without quadratic cost.}
    \label{fig:hyena}
\end{figure}

\begin{figure}
    \centering
    \includegraphics[width=0.8\textwidth]{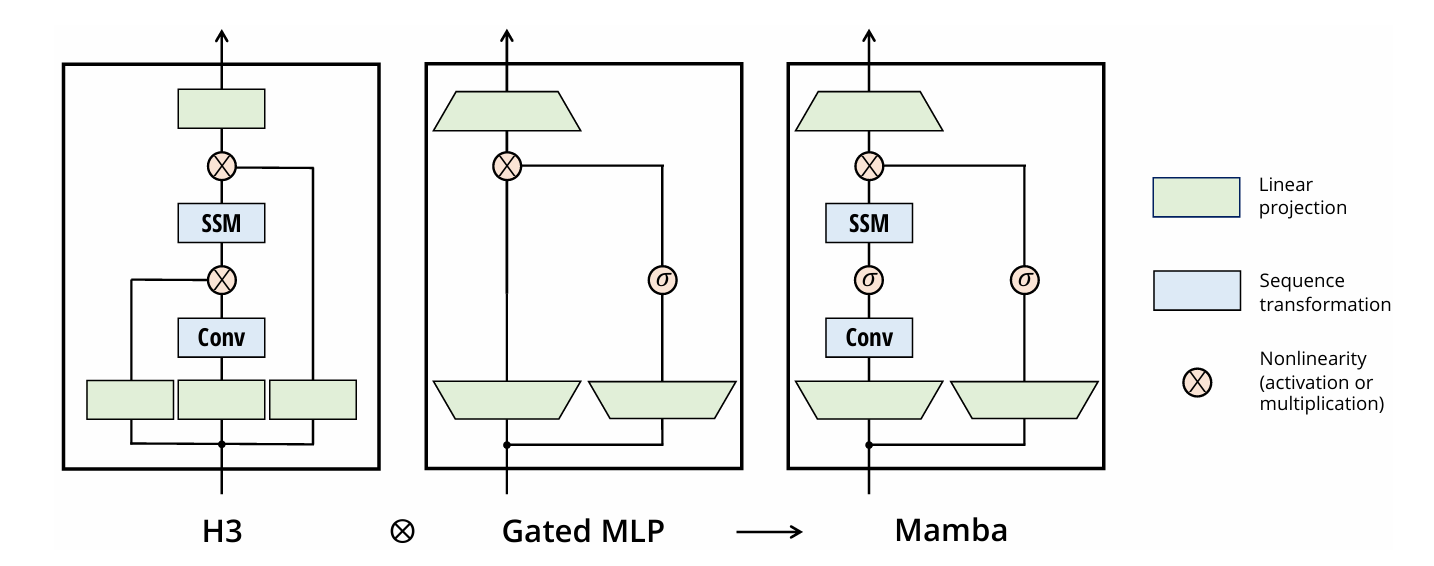}
    \caption{Mamba architecture~\cite{gu2023mamba}. Uses state-space sequence modeling (SSM) with gating and convolution to replace self-attention.}
    \label{fig:mamba}
\end{figure}

We evaluate four encoder-only architectures with matched parameter budgets:

\begin{itemize}
    \item \textbf{Baseline Transformer:} GPT2-style decoder-only transformer with causal self-attention~\cite{vaswani2017attention}.
    \item \textbf{FlashAttention Transformer:} Variant with FlashAttention kernels~\cite{dao2022flashattention} for IO-aware optimized attention (See Figure \ref{fig:flashattention}).
    \item \textbf{Hyena Transformer:} Replaces self-attention with Hyena operators~\cite{poli2023hyena}, using convolutional modulation mechanisms (See Figure \ref{fig:hyena}).
    \item \textbf{Mamba:} Selective state space model using implicit continuous-time recurrence~\cite{gu2023mamba} (See Figure \ref{fig:mamba}). 
\end{itemize}

All models are trained from scratch and evaluated under the same context-query formulation.

\subsection{Training Procedure}

We train all models using the squared error loss between predicted and target query outputs:
\[
\mathcal{L} = \frac{1}{B} \sum_{b=1}^B \left(f_\theta(x_{T+1}^{(b)}) - y_{T+1}^{(b)}\right)^2.
\]

\subsection{Curriculum Learning}

To improve convergence and stability, we adopt curriculum learning~\cite{bengio2009curriculum,wu2020whendo,elman1993starting}. During training, tasks are sampled from small dimensions, gradually increasing task complexity as training proceeds. This allows faster convergence, especially for difficult function classes like chaotic dynamics.

\subsection{Evaluation Criteria}
\paragraph{Baseline Estimators.} 
To assess the generalization and efficiency of each model, we compare their in-context learning (ICL) performance against a set of reference estimators: \textbf{zero estimator}, \textbf{least squares}, \textbf{3-nearest neighbor}, and \textbf{averaging}. These baselines are selected to span a range of statistical and algorithmic properties, offering insight into both trivial and non-trivial learning behaviors, in line with established ICL evaluation protocols~\citep{garg2023transformerslearnincontextcase}.

\begin{itemize}
    \item \textbf{Zero Estimator.} This predicts a constant output of zero regardless of the input:
    \[
    M(P) = 0.
    \]
    As a trivial baseline, it sets a lower bound for model performance and ensures that any positive result reflects non-trivial learning.

    \item \textbf{Least Squares Estimator.} This is the minimum-norm solution to the linear regression problem, optimal under Gaussian noise. Given a prompt \( P = \{(x_1, y_1), \ldots, (x_k, y_k), x_{\text{query}}\} \), define \( X \in \mathbb{R}^{k \times d} \) as the matrix with rows \( x_i^\top \), and \( y \in \mathbb{R}^k \) as the output vector. The prediction is:
    \[
    \hat{w}^\top = X^+ y, \quad M(P) = \hat{w}^\top x_{\text{query}},
    \]
    where \( X^+ \) denotes the Moore-Penrose pseudoinverse of \( X \). This serves as a gold-standard baseline for linear tasks.

    \item \textbf{3-Nearest Neighbor Estimator.} This method averages the outputs corresponding to the three inputs closest to \( x_{\text{query}} \) in Euclidean distance. Let \( S \subseteq \{1, \ldots, k\} \) be the indices of the 3 nearest neighbors. Then,
    \[
    M(P) = \frac{1}{|S|} \sum_{i \in S} y_i.
    \]
    This non-parametric estimator evaluates whether models can exploit local geometric structure in the context, especially for tasks with non-linear dependencies.

    \item \textbf{Averaging Estimator.} This approach predicts using the average of all context outputs weighted by inputs:
    \[
    \hat{w} = \frac{1}{k} \sum_{i=1}^k x_i y_i, \quad M(P) = \hat{w}^\top x_{\text{query}}.
    \]
    While not optimal, this estimator is consistent under standard assumptions (e.g., \( x_i \sim \mathcal{N}(0, I_d) \)) and avoids matrix inversion, making it more likely to be captured by simple ICL mechanisms.
\end{itemize}
(2023)~\citep{garg2023transformerslearnincontextcase}, allowing us to focus on their comparative roles.

Performance is assessed using the following criteria:

\begin{itemize}
    \item Mean Squared Error (MSE) over test prompts
    
    The Mean Squared Error (MSE) is defined as:
\[
\text{MSE} = \frac{1}{n} \sum_{i=1}^{n} (y_i - \hat{y}_i)^2
\]
where \( y_i \) is the actual value, \( \hat{y}_i \) is the predicted value, and \( n \) is the number of observations.
    \item Generalization to new task instances from each function family
    \item Robustness under context length variation and input noise
    \item Scaling behavior as context length $T$ increases
\end{itemize}

Model parameters are not updated at the test time. In all cases, the model must extrapolate in-context based solely on the prompt.

\section{Experiments}

\subsection{Task and Dataset}

Following the experimental design of~\cite{garg2023transformerslearnincontextcase}, we evaluate in-context learning (ICL) capabilities within a controlled synthetic framework. In this setup, models learn functions from a class $F$ using prompt-based adaptation—\textbf{without any explicit parameter updates}. Learning performance is determined by the model's ability to generalize to new inputs and functions solely based on in-context examples.

\paragraph{Function Sampling.}
Each episode begins with sampling a function $f \sim D_F$, and a sequence of $n$ inputs $x_1, x_2, \dots, x_n \sim D_X$, where both $D_F$ and $D_X$ are pre-defined distributions over function classes and input domains, respectively.

\paragraph{Prompt Construction.}
A prompt is constructed using the first $k$ input-output pairs and a query input:
\[
P = (x_1, f(x_1), \dots, x_k, f(x_k), x_{k+1}).
\]
The model is then tasked with predicting $f(x_{k+1})$.

\paragraph{Evaluation Objective.}
This formulation allows direct measurement of a model's intrinsic in-context learning capabilities, i.e., its ability to \textbf{adapt to new functions} using only context, not gradient-based learning. The loss is computed via \textbf{Mean Squared Error (MSE)} on the model's prediction of $f(x_{k+1})$.

\paragraph{Function Classes.}
We focus on two complex function classes to evaluate model generalization:
\begin{itemize}
    \item \textbf{Gaussian Kernel Regression:} Each function is a sum of 20 Gaussian kernels with bandwidth $\sigma = 1.5$, where both the kernel centers and weights are sampled from \( \mathcal{N}(0, 1) \). Outputs are perturbed by Gaussian noise with standard deviation 0.1.
    \item \textbf{Nonlinear Dynamical Systems:} Functions are defined by polynomial dynamics up to degree 3, with coefficients sampled from \( \mathcal{N}(0, 1) \). These functions are deterministic and emphasize recursive temporal dependencies.
\end{itemize}

\paragraph{Input Distribution and Dimensions.}
Inputs are sampled uniformly from the domain \( \text{Unif}([-1, 1]^d) \). We evaluate across multiple input dimensions \( d \in \{1, 10, 50, 100\} \), to test both low- and high-dimensional generalization.

\paragraph{Training Details.}
\begin{itemize}
    \item \textbf{Prompt length:} $k = 20$ unless stated otherwise.
    \item \textbf{Batch size:} 64 episodes per batch.
    \item \textbf{Data generation:} Performed on-the-fly to ensure function diversity.
    \item \textbf{Random seeds:} Fixed for training, varied for testing for robust generalization measurement.
\end{itemize}

\subsection{Model Structure and Variants}

We primarily adopt a \textbf{decoder-only transformer} architecture inspired by GPT-2~\cite{garg2023transformerslearnincontextcase}, and evaluate several architectural variants to understand how design choices affect ICL performance.

\paragraph{Base Transformer Architecture.}
\begin{itemize}
    \item \textbf{Layers:} 12 Transformer decoder layers
    \item \textbf{Attention heads:} 8
    \item \textbf{Embedding dimension:} 256
    \item \textbf{Input representation:} Each scalar input and output is independently projected into the embedding space using separate \textbf{learnable linear layers}. The output $f(x_i)$ is zero-padded to match input dimensionality before projection.
    \item \textbf{Prediction target:} The model predicts the embedding of $f(x_{k+1})$ using previous $k$ examples in context.
\end{itemize}

\paragraph{Alternative Architectures.}
We evaluate additional models with alternative inductive biases or improved efficiency:
\begin{itemize}
    \item \textbf{FlashAttention}~\cite{dao2022flashattention}: A memory- and compute-efficient implementation of attention, allowing faster training and inference without sacrificing accuracy.
    \item \textbf{Hyena}~\cite{poli2023hyena}: A convolutional architecture designed for long-range dependencies, replacing attention with structured convolutions.
    \item \textbf{Mamba}~\cite{gu2023mamba}: A \textbf{state space model} that dispenses with attention entirely. In our configuration:
    \begin{itemize}
        \item \textbf{Number of layers:} 24
        \item \textbf{Attention heads:} 0 (fully attention-free)
        \item \textbf{Embedding size:} 256
    \end{itemize}
\end{itemize}

This diverse set of architectures enables a comparative study of how different model structures handle in-context adaptation tasks, especially in the presence of nontrivial function structure and noise.

\subsection{Model Training}
\subsubsection{Training Details}

We train all models under a unified objective using the mean squared error (MSE) loss, following the protocol of~\cite{garg2023transformerslearnincontextcase}. All experiments are conducted on NVIDIA RTX 4090 GPUs. Models are trained from scratch with randomly initialized parameters, updated via AdamW optimizer using gradient descent.

We explore batch sizes from ${64, 128}$, selecting the optimal size based on model memory requirements and convergence speed. Each model is trained for 50k steps by default, unless otherwise stated. To ensure diversity and minimize memorization, each training batch consists of freshly sampled synthetic functions and inputs (on-the-fly generation). The training dataset comprises 10k unique function instances, with an additional 1k examples reserved for validation and testing.

For the learning rate, we choose from ${1 \times 10^{-4}, 5 \times 10^{-5}}$ depending on the model type and task complexity. In general, larger learning rates are used for smaller or more stable architectures (e.g., vanilla Transformers), while more sensitive or deeper architectures (e.g., Mamba) benefit from a smaller learning rate. When applicable, we employ a cosine learning rate schedule with a linear warm-up of 3k steps to stabilize early training and allow better convergence in later phases.

For the nonlinear dynamical system task, due to its recursive and temporally entangled nature, we increase the number of training steps (up to 100k in some configurations) and apply early stopping based on validation loss to prevent overfitting and ensure generalization.

\subsubsection{Curriculum Learning}
\label{subsec:curriculum_learning}

To improve training stability and efficiency—especially in high-dimensional settings—we adopt a \textbf{curriculum learning} strategy that gradually increases task difficulty along two axes: input subspace dimension and prompt length.

\paragraph{Initialization.}
For both function classes, training begins with inputs sampled from a 5-dimensional subspace of the full input space, with remaining coordinates zero-padded. We also initialize with short prompts to ease early-stage sequence modeling:

\begin{itemize}
    \item \textbf{Gaussian kernel regression (linear-like):} prompt length = 11 (i.e., 11 input-output pairs).
    \item \textbf{Nonlinear dynamical systems:} prompt length = 26, following~\cite{garg2023transformerslearnincontextcase}, to better capture recursive structure.
\end{itemize}

\paragraph{Progression.}
Every 2k training steps, we:

\begin{itemize}
    \item Increase the input subspace dimension by 1;
    \item Extend the prompt length by:
    \begin{itemize}
        \item +2 tokens for Gaussian kernel regression;
        \item +5 tokens for nonlinear dynamics.
    \end{itemize}
\end{itemize}

This progression continues until the full input dimensionality ($d = 20$) and target prompt lengths (41 and 101 respectively) are reached.

\paragraph{Effect.}
This curriculum empirically accelerates convergence, mitigates early-stage instability, and improves generalization in high-dimensional and long-context settings.

\section{Results}
\label{others}

\subsection{Tasks}
\begin{figure}
  \centering
  \begin{subfigure}[t]{0.45\textwidth}
    \centering
    \includegraphics[width=\linewidth]{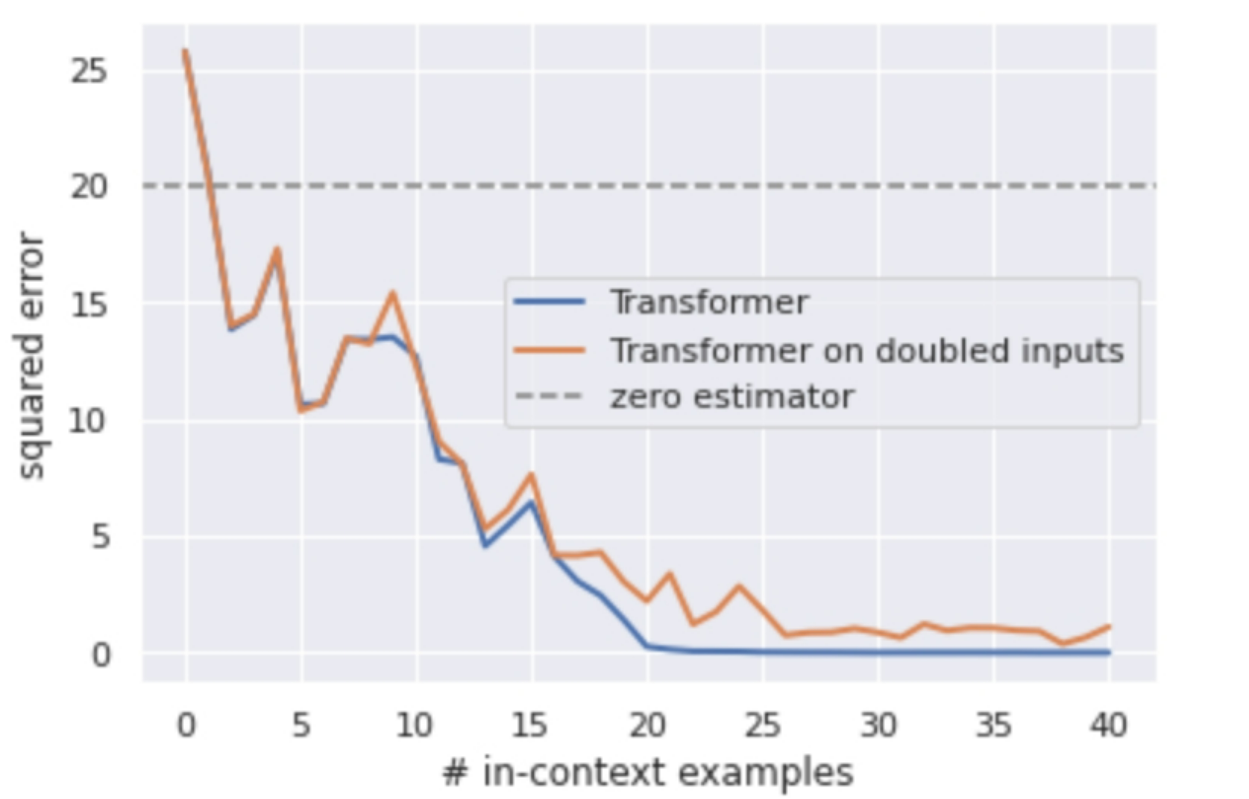}
    \caption{Performance of linear regression tasks with GPT-2.}
    \label{fig:sub1}
  \end{subfigure}
  \hfill
  \begin{subfigure}[t]{0.45\textwidth}
    \centering
    \includegraphics[width=\linewidth]{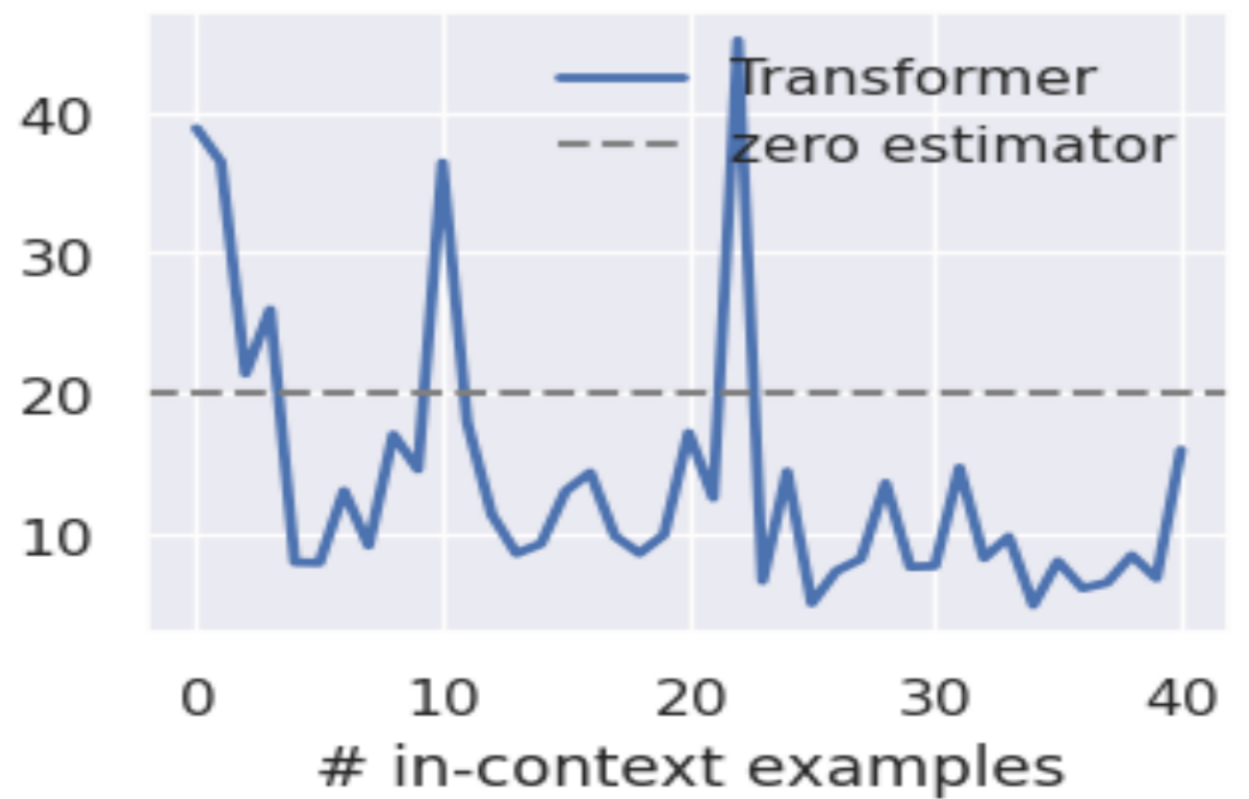}
    \caption{GPT-2 vs. zero estimator on Gaussian kernel regression.}
    \label{fig:sub2}
  \end{subfigure}

  \vspace{0.3cm}

  \begin{subfigure}[t]{0.45\textwidth}
    \centering
    \includegraphics[width=\linewidth]{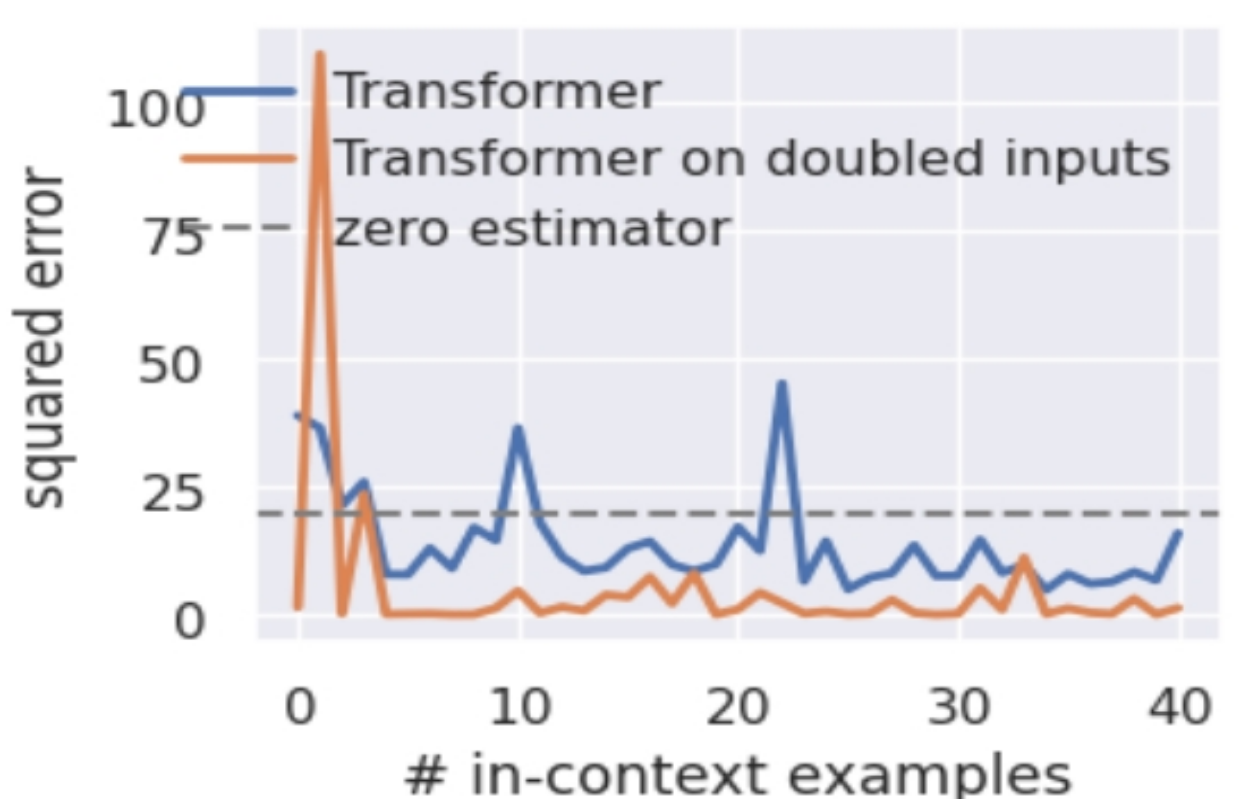}
    \caption{Standard vs. input-doubled GPT-2 on Gaussian regression.}
    \label{fig:sub3}
  \end{subfigure}
  \hfill
  \begin{subfigure}[t]{0.45\textwidth}
    \centering
    \includegraphics[width=\linewidth]{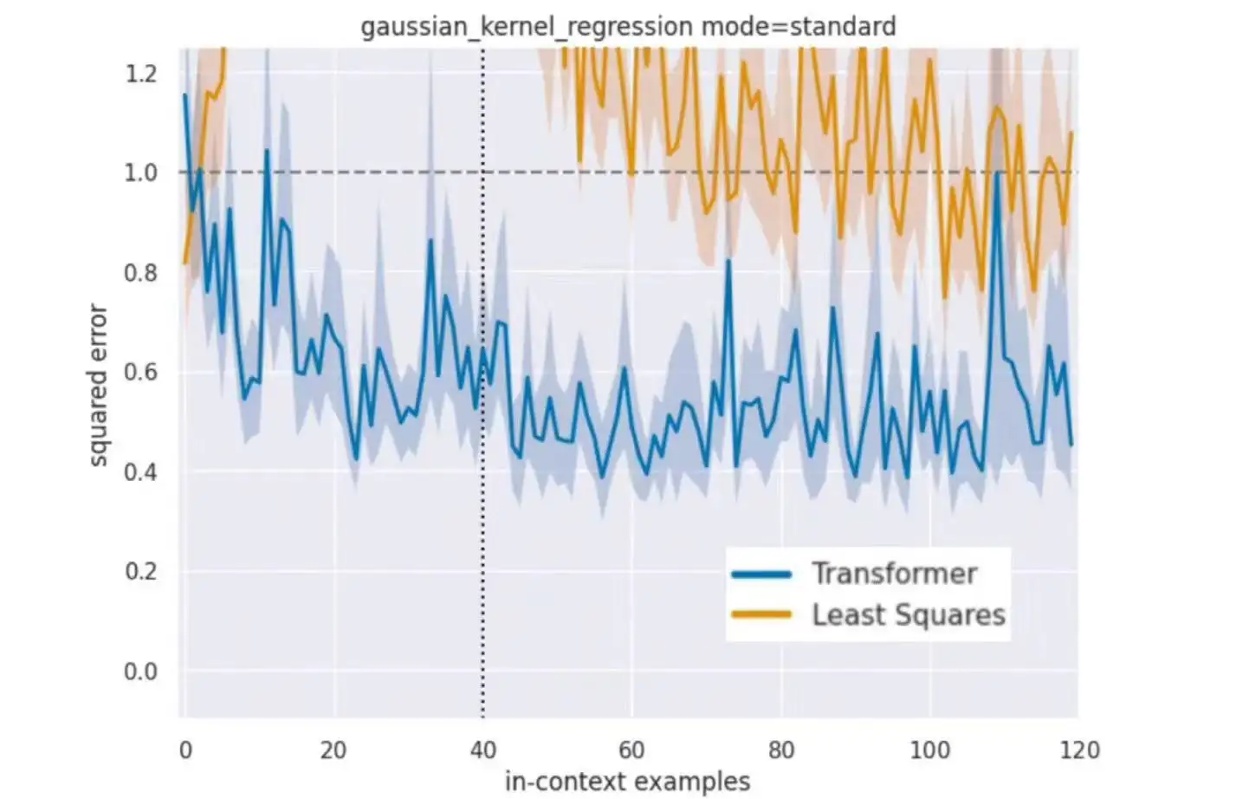}
    \caption{Effect of increasing in-context examples on GPT-2.}
    \label{fig:sub4}
  \end{subfigure}

  \caption{Summary of GPT-2 results on linear and Gaussian kernel regression tasks.}
  \label{fig:gpt2_summary}
\end{figure}

\begin{figure}[htbp]
  \centering

  \begin{subfigure}[t]{0.3\textwidth}
    \centering
    \includegraphics[width=\linewidth]{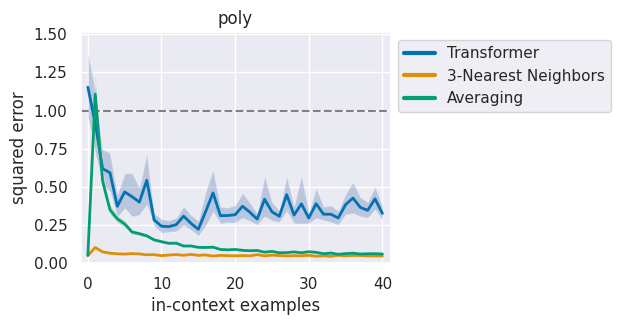}
    \caption{polynomial functions}
    \label{fig:sub5}
  \end{subfigure}
  \hfill
  \begin{subfigure}[t]{0.3\textwidth}
    \centering
    \includegraphics[width=\linewidth]{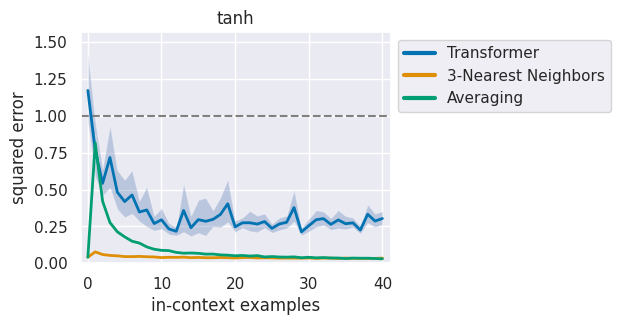}
    \caption{tanh functions}
    \label{fig:sub6}
  \end{subfigure}
  \hfill
  \begin{subfigure}[t]{0.3\textwidth}
    \centering
    \includegraphics[width=\linewidth]{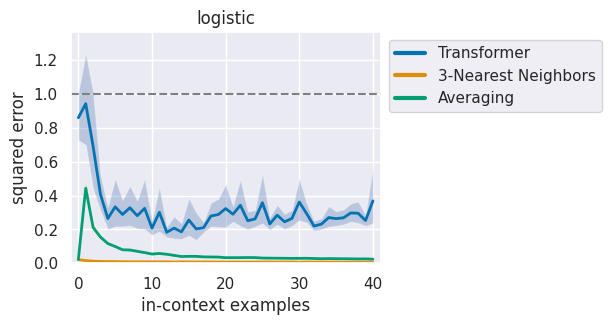}
    \caption{logistic functions}
    \label{fig:sub7}
  \end{subfigure}

  \vspace{0.5cm}  

  \begin{subfigure}[t]{0.3\textwidth}
    \centering
    \includegraphics[width=\linewidth]{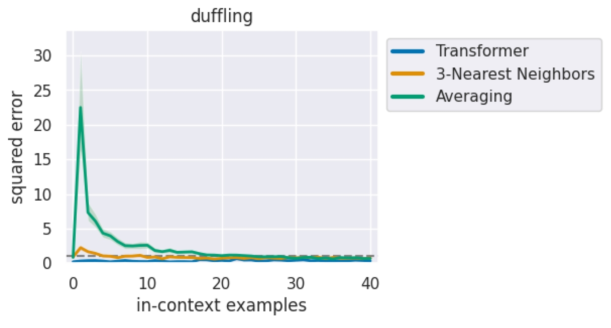}
    \caption{duffling functions}
    \label{fig:sub8}
  \end{subfigure}
  \hfill
  \begin{subfigure}[t]{0.3\textwidth}
    \centering
    \includegraphics[width=\linewidth]{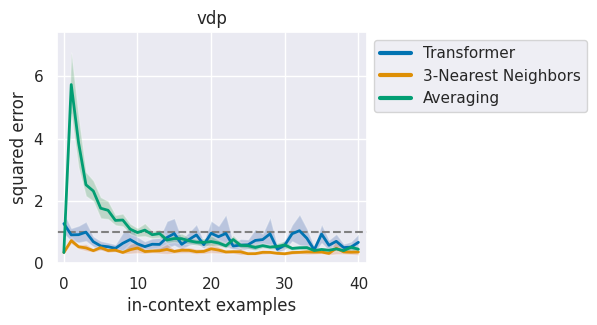}
    \caption{vdp functions}
    \label{fig:sub9}
  \end{subfigure}
  \hfill
  \begin{subfigure}[t]{0.3\textwidth}
    \centering
    \includegraphics[width=\linewidth]{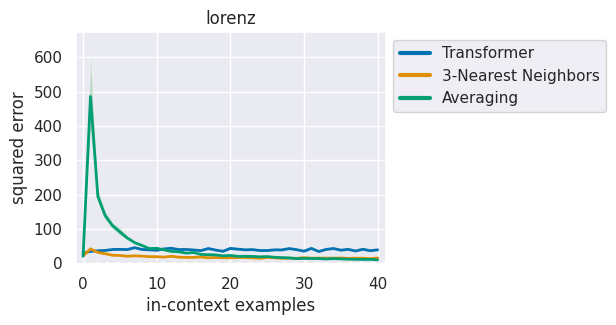}
    \caption{lorenz functions}
    \label{fig:sub10}
  \end{subfigure}

  \caption{Results of Nonlinear Dynamics Trained with GPT-2 Architecture}
  \label{fig:six_results}
\end{figure}

\textbf{Gaussian Kernel Regression}
Based on GPT 2 architecture, the performance of Gaussian kernel regression task(\ref{fig:sub2}) exhibits a fluctuating pattern with an overall mild downward trend, yet lacks clear stability compared with the linear regression task(\ref{fig:sub1}). Despite outperforming the zero estimator on average, the Transformer displays noticeable instability, with several spikes exceeding the baseline error. This suggests that the model’s ability to utilize in-context examples effectively is limited in this setting. However, applying doubled inputs(\ref{fig:sub3}) significantly reduces the squared error and yields a more stable performance compared to the standard Transformer. Continuing to grow the amount of in-context examples, the model achieved better results on Gaussian kernel regression, showing a more stable trend of deceasing(\ref{fig:sub4}).
\paragraph{\textbf{Nonlinear Dynamics}}
During the training process of nonlinear dynamical systems, the loss generally increases as the dimensionality, the number of data points, and task difficulty grow. However, within certain intervals, the loss decreases, indicating that the model benefits from gradually increasing complexity. 
Among the evaluated dynamical systems, they all showed a trend of decreasing.(\ref{fig:six_results}) Functions such as \textit{tanh} and \textit{poly} exhibit fast and smooth convergence as the number of in-context examples increases, as they have relatively lower complexity and higher compatibility with in-context learning. In contrast, the \textit{Lorenz} system shows a significantly higher initial error and slower convergence, which is consistent with its known chaotic behavior and intrinsic complexity. The \textit{duffling} system demonstrates a sharp decline in error with only a few examples, highlighting its strong sensitivity to the number of in-context samples. Meanwhile, \textit{logistic} and \textit{vdp} systems present intermediate patterns in both convergence speed and final error, reflecting their moderate learning difficulty. 

Both tasks are more complex than linear regression, and although their results are less ideal, they still demonstrate that the model has, to some extent, acquired knowledge of these functions through in-context learning. 

\subsection{Attention Implementation / Architecture}
\begin{figure}[htbp]
  \centering

  \begin{subfigure}[t]{0.45\textwidth}
    \includegraphics[width=\linewidth]{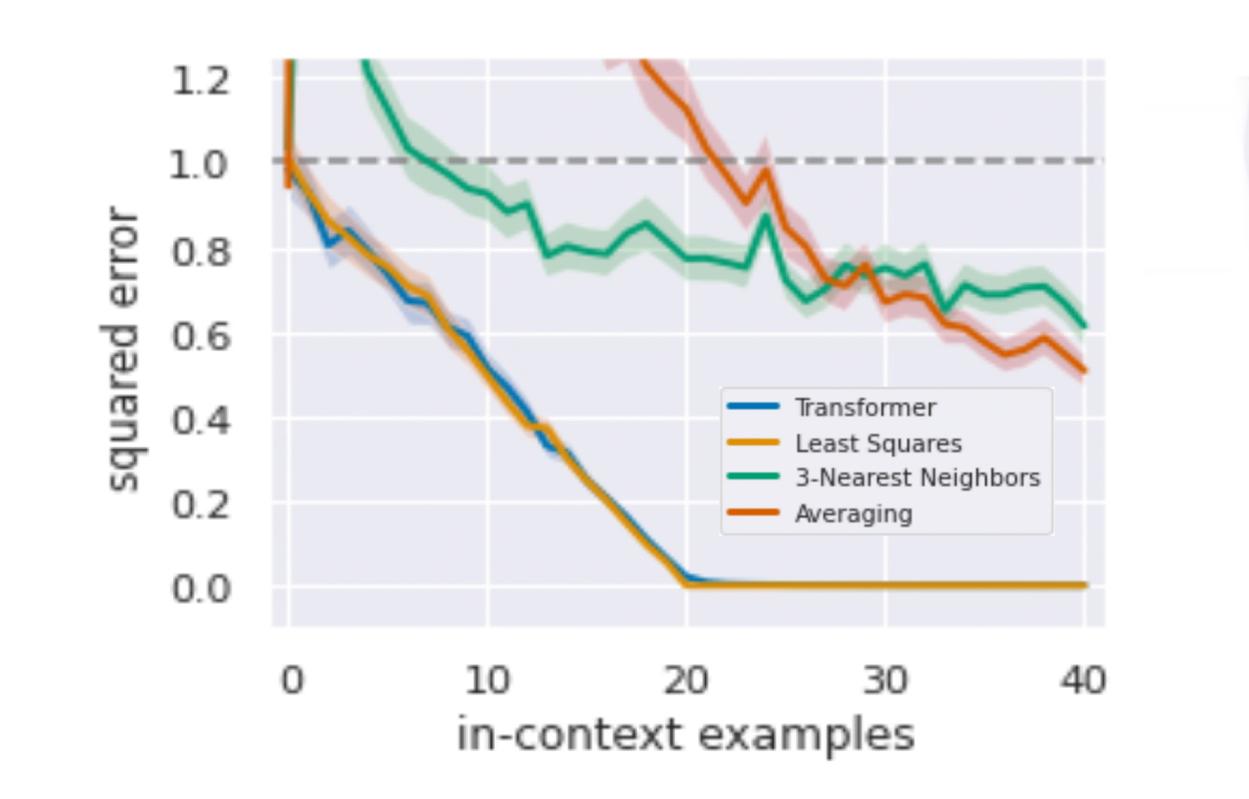}
    \caption{Transformer}
    \label{fig:sub11}
  \end{subfigure}
  \hfill
  \begin{subfigure}[t]{0.45\textwidth}
    \includegraphics[width=\linewidth]{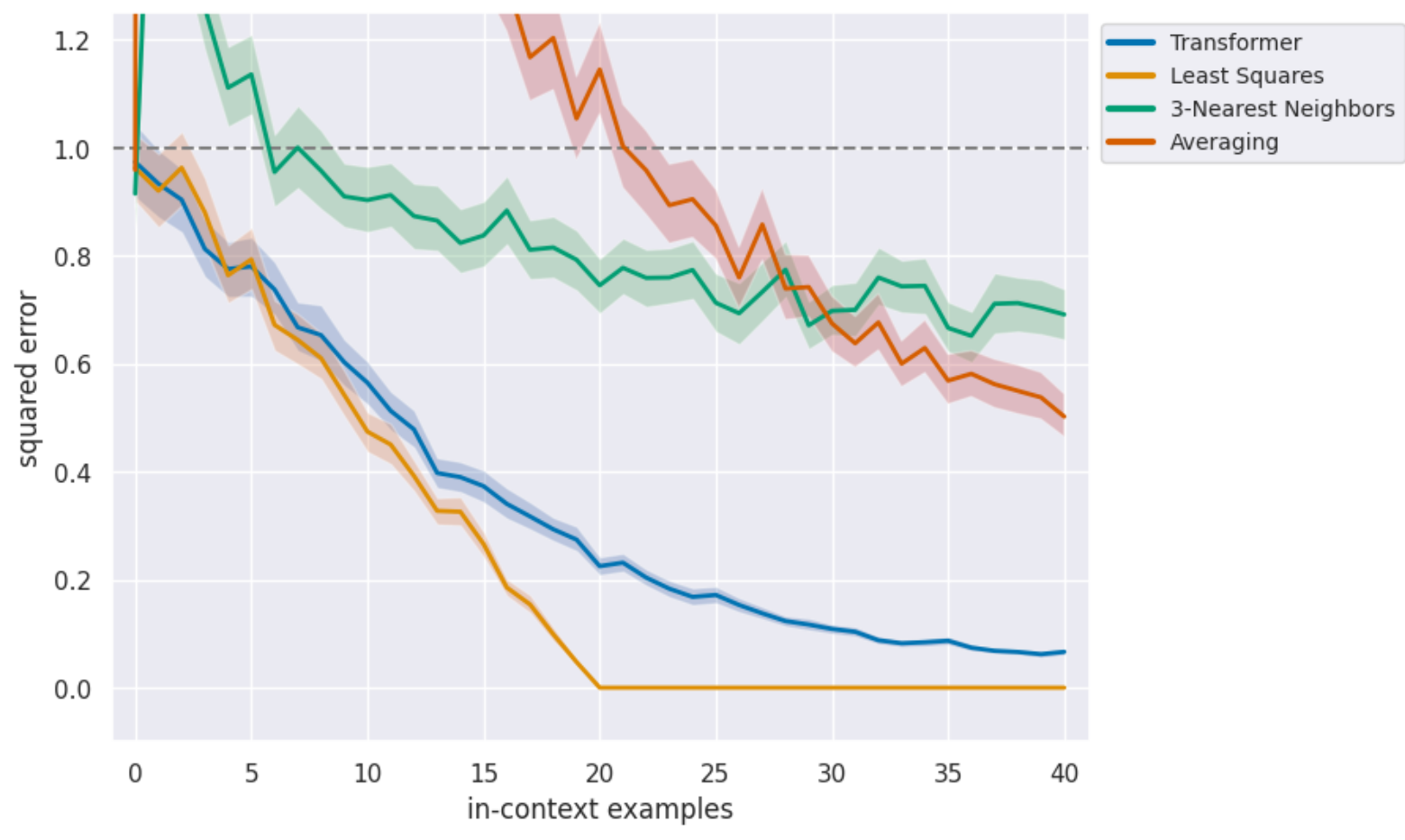}
    \caption{Transformer with Hyena}
    \label{fig:sub12}
  \end{subfigure}

  \vspace{0.5cm}  

  \begin{subfigure}[t]{0.45\textwidth}
    \includegraphics[width=0.95\linewidth, height=0.6\linewidth]{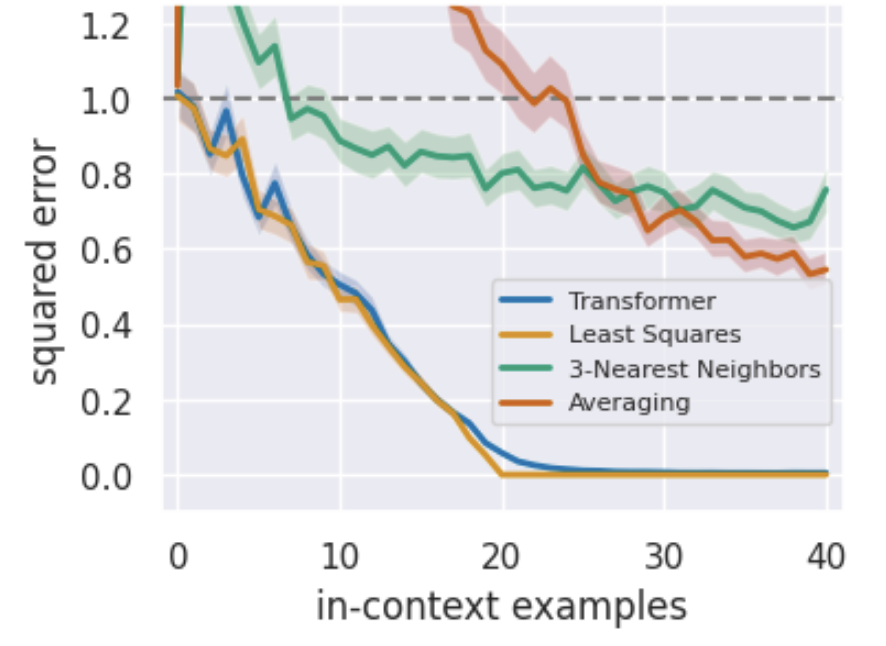}
    \caption{Transformer with Flash Attention}
    \label{fig:sub13}
  \end{subfigure}
  \hfill
  \begin{subfigure}[t]{0.45\textwidth}
    \includegraphics[width=\linewidth]{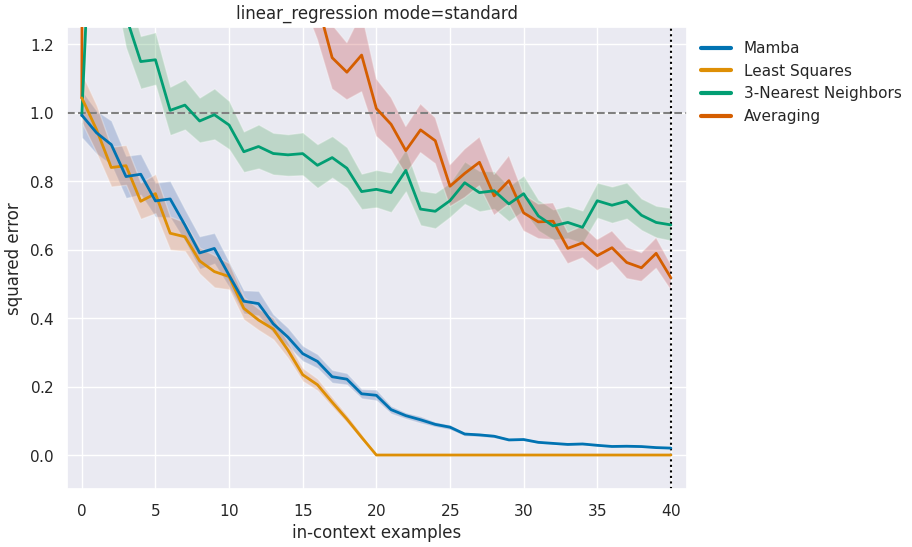}
    \caption{Mamba}
    \label{fig:sub14}
  \end{subfigure}

  \caption{Results of 4 architectures on linear regression task}
  \label{fig:architecture_2x2}
\end{figure}

\paragraph{Hyena}
We compare the performance of a standard Transformer baseline and a Transformer augmented with Hyena(\ref{fig:sub12}) filters on the same linear regression task. Although the Hyena-augmented model starts with higher initial error and greater early-stage variability, it exhibits a consistent downward trend and eventually achieves comparable performance. This progression indicates that the model is actively learning from context, not merely memorizing, and that the Hyena filters offer sufficient representational capacity for in-context learning despite their non-attentional nature.
\paragraph{\textbf{Flash Attention}}
Evaluating the GPT 2 model with flash attention on the linear regression task(\ref{fig:sub13}), while the Transformer equipped with Flash Attention achieves results that are generally consistent with the baseline Transformer, its performance is marginally lower. The model performs poorly on Gaussian kernel regression, with an error peak around 20 examples, while it shows lower and decreasing errors on Nonlinear Dynamics. 
\paragraph{\textbf{Mamba}}
In the linear regression task, Mamba(\ref{fig:sub14}) shows a consistent reduction in error with more in-context examples, outperforming the zero estimator and approaching the performance of the Transformer. This indicates that the model is not guessing but indeed learning from context. 
In comparison, the results on Gaussian kernel regression are moderate, better than Least Squares but not very well, while performance on Nonlinear Dynamics is acceptable despite some initial fluctuations. 

Among the four implementations, the standard Transformer exhibits the most stable learning behavior, with smooth error reduction and strong final convergence. Mamba shows consistent and reliable performance throughout training, with error curves closely aligned with the Least Squares baseline, albeit with a slower learning rate in the early stages. Hyena demonstrates efficient learning and strong accuracy, though its initial performance can be more sensitive to sample size. Flash Attention achieves rapid convergence as the number of in-context examples increases, but exhibits larger fluctuations in the early phase, especially under limited data conditions.

\section{Discussion}

\subsection{Architectural Adaptation on Function Properties}

\begin{figure}[htbp]
  \centering
  \begin{minipage}[t]{0.48\textwidth}
    \centering
    \includegraphics[height=4cm]{result/nonlinear_dynamics/nonlinear_dynamics_poly.jpg}
    \caption*{(a) Transformer}
    \label{fig:sub15}
  \end{minipage}
  \hfill
  \begin{minipage}[t]{0.48\textwidth}
    \centering
    \begin{tikzpicture}[spy using outlines={rectangle, magnification=2.5, size=2.5cm, connect spies}]
      \node {\includegraphics[height=3.5cm]{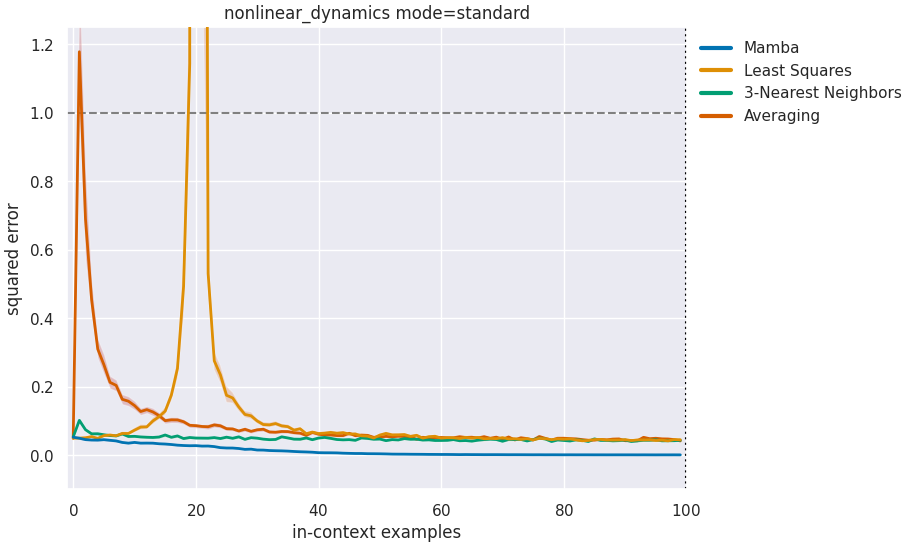}};
      \spy on (-2,-0.8) in node [right] at (1,2.8);
    \end{tikzpicture}
    \caption*{(b) Mamba}
    \label{fig:sub16}
  \end{minipage}

  \vspace{0.3cm}
  \caption{Comparison between the capability on nonlinear dynamics of Transformer and Mamba}
  \label{fig:architecture_1x2}
\end{figure}

Our comparative study across four model implementations --- GPT-2-style Transformers, FlashAttention-enhanced Transformers, Hyena, and Mamba---reveals that \textbf{model architecture strongly biases performance across different function families} in in-context learning (ICL). Transformer-based models exhibit relatively stable performance across all evaluated tasks, reflecting their general-purpose inductive bias and full-context attention mechanism~\cite{vaswani2017attention}. However, they are constrained by quadratic scaling in compute and limited context lengths, even with optimizations like FlashAttention~\cite{dao2022flashattention}. In contrast, \textbf{Mamba excels in tasks involving recursive structure and temporal dependencies}, such as nonlinear dynamics(\ref{fig:architecture_1x2}), achieving strong performance at significantly lower computational cost. This advantage stems from Mamba’s structured state-space design~\cite{gu2023mamba}, which enables efficient sequential reasoning and localized integration of information without full prompt attention. Hyena~\cite{poli2023hyena} falls between these extremes, leveraging long-range convolutions, but its hybrid nature may diffuse its inductive alignment with any particular function class. These findings support the view that \textbf{architectural alignment with the target function's structure is critical to ICL success}, especially for tasks with algorithmic or dynamical properties.

\subsection{Localization Effect Caused by Gaussian Kernel}
Initial experiments on Gaussian kernel regression revealed that \textbf{naively applying a bare Gaussian kernel formulation leads to trivial solutions} (see Appendix Figure~\ref{fig:kernel_locality},~\ref{fig:readout_loss}), with models achieving near-zero evaluation error regardless of training. This occurs because such kernels act as local interpolators: when support points are densely clustered, the model can exploit local smoothing to produce accurate outputs without needing to extract or generalize from the structure of in-context examples.

To counteract this, we reframed the task by applying a \textbf{linear readout layer on top of the Gaussian similarity features}, turning the model’s objective into one of learning weighted combinations of localized kernels. While this adjustment made the task more representative and challenging---restoring error curves to expected behavior (Appendix Figure~\ref{fig:readout_loss})---it also introduced high \textbf{variance across evaluation runs}. We attribute this to the \textbf{sensitivity of Gaussian kernels to input distribution geometry}, particularly under small bandwidths or uneven spacing of support points. These results suggest that kernel-based ICL tasks must be carefully framed to balance local smoothness with global compositional reasoning.

\subsection{Exploitation on Nonlinear Terms for Geometric Separability}
In robustness experiments inspired by~\cite{garg2022can}, we evaluated model behavior under doubled input domains. Surprisingly, models often \textbf{performed better when the input range was expanded}, especially on nonlinear dynamics tasks. We interpret this phenomenon through the lens of \textbf{geometric separability in representation space} (Appendix Figure~\ref{fig:tsne_latents}). When input $x$-values are confined to $[-1, 1]$, higher-order terms such as $x^2$ and $x^3$ exhibit minimal variation, making it difficult for the model to distinguish between support and query points. Doubling the input domain to $[-2, 2]$ amplifies local variation, especially in nonlinear terms, thereby \textbf{enhancing representational contrast}.

Additionally, when outputs are normalized post-scaling, the transformation effectively injects sharper curvatures and larger gradients into the same output range. These changes make derivative patterns more salient and \textbf{easier to detect by local mechanisms} like Mamba’s convolutional state updates or attention weights in Transformers. This behavior is visualized in Appendix Figure~\ref{fig:fx_scaling}, showing amplified curvature and steeper slopes for scaled inputs. In this sense, input scaling can serve as a form of \textbf{implicit feature amplification}, improving sample efficiency and generalization on complex nonlinear functions.

\subsection{Mechanism Behind Curriculum Alignment}
We also identify a deeper structure underlying the curriculum learning strategy proposed by~\cite{garg2022can}. Their method incrementally increases the input dimension and context length in synchronized stages. Upon analysis, we observe that the \textbf{context length scaling ratio differs based on the complexity of the target function class}: for linear regression tasks, the context length grows modestly to $2d + 1$, while for more expressive function families such as decision trees and two-layer neural networks, it expands more aggressively to $5d + 1$ (Appendix Figure~\ref{fig:curriculum_schedule}).

This scaling ensures that more complex models observe sufficiently rich prompts to recover global structure, without overshooting the optimization budget. We further connect this to \textbf{gradient starvation and symmetry breaking in non-curriculum training}: starting with high-dimensional prompts leads to negligible gradient signals due to orthogonality and uniform input influence, causing models to stagnate until a mechanism is discovered. In contrast, curriculum learning offers a warm start in low-dimensional settings, progressively expanding task complexity while preserving training signal strength. This results in \textbf{earlier mechanism discovery and faster convergence}, as confirmed by training comparisons in Garg et al. (2023)~\citep{garg2023transformerslearnincontextcase}.

\section{Conclusion}

In this work, we presented a evaluation framework for studying in-context learning (ICL) behaviors across a diverse set of function families and model architectures. Our experiments demonstrate that the architectural choices can have a rather strong impact on ICL performance, particularly under tasks with recursive or nonlinear temporal dependencies. We find that Mamba, a structured state space model, excels on nonlinear dynamical systems, while Transformers exhibit robust generality. Furthermore, we reveal the subtle phenomena such as the localization bias in Gaussian kernels, implicit feature amplification through input scaling, and convergence benefits from curriculum learning.

There are a few directions that can be explored next. First, we saw that different model architectures behave differently depending on the type of function they’re working with. The function types may be broken down more carefully to investigate which models are best suited for which class, which could help us better understand the kinds of problems each model is naturally good at. Since Mamba seems to do well with time-related tasks, a natural step is to try mixing it with Transformers to build a model that handles both long-range and step-by-step reasoning.

Also, we noticed that when we made the input range larger, the models actually learned better, especially for non-linear tasks. This might be because the differences between input values became more noticeable, making it easier for the model to pick up patterns. This can be studied more carefully, with smarter ways coming up to scale or reshape the inputs so that the important features stand out more and learning becomes easier.

Despite these findings, our study has several limitations. First, the evaluation framework primarily focuses on synthetic tasks with well-defined function families, such as polynomial (\ref{eq:polynomial}) and chaotic systems. While these tasks provide controlled settings to study ICL, they may not fully capture the complexity of real-world applications, where data distributions are often noisier and less structured. Second, the curriculum learning strategy (\ref{subsec:curriculum_learning}) was tailored to specific dimensional and prompt length progressions, which may not generalize optimally across all model architectures or task types. Finally, our analysis of architectural performance, while comprehensive, is limited by the computational resources available, restricting the scale of models and the breadth of hyperparameter tuning. These constraints suggest caution when extrapolating our findings to larger models or diverse domains, motivating further investigation in the directions outlined above.


\bibliographystyle{plain}
\bibliography{refs}


\newpage
\appendix

\section{OOD: Out-of-Distribution Experiments}
This section of appendix is a supplement to the result of out-of-distribution experiments with abundant visualization.




\subsection{Transformer with Flash Attention}
\begin{figure}[htbp]
  \centering

  \begin{subfigure}[t]{0.3\textwidth}
    \centering
    \includegraphics[width=\linewidth]{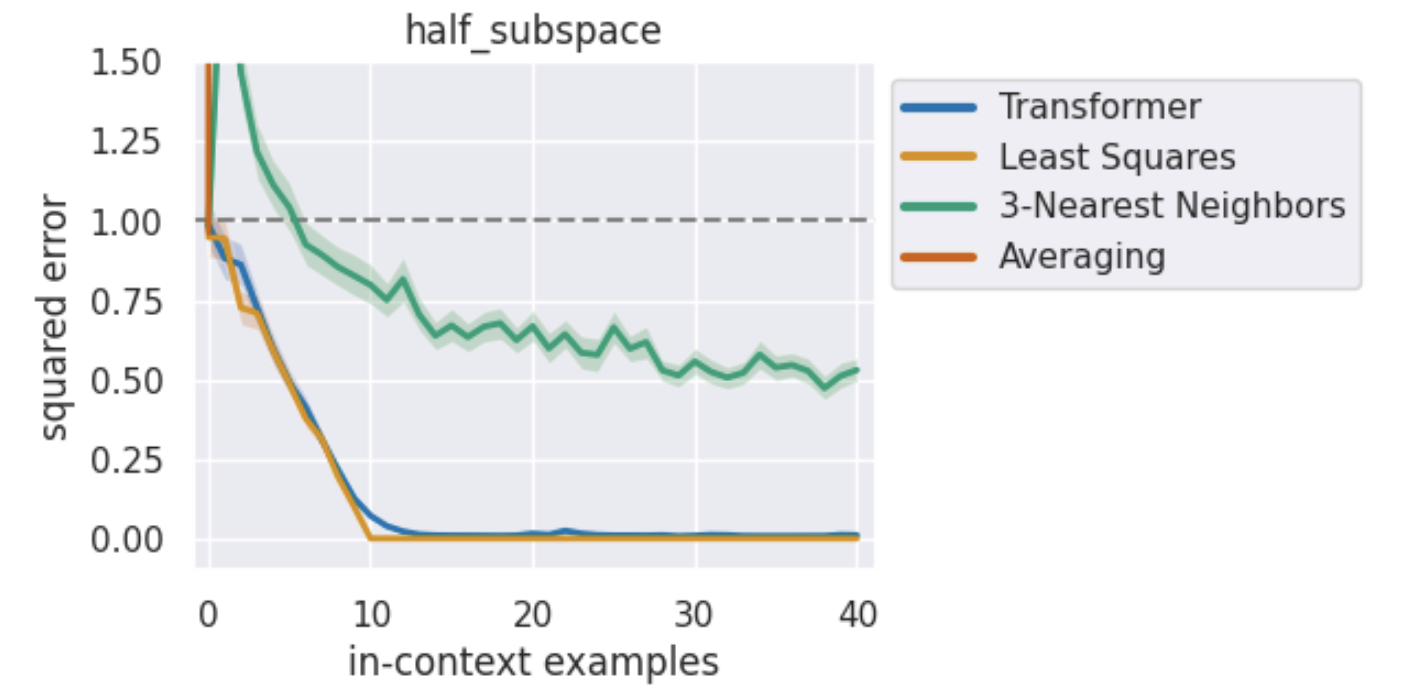}
    \caption{Half-subspace}
    \label{fig:sub17}
  \end{subfigure}
  \hfill
  \begin{subfigure}[t]{0.3\textwidth}
    \centering
    \includegraphics[width=\linewidth]{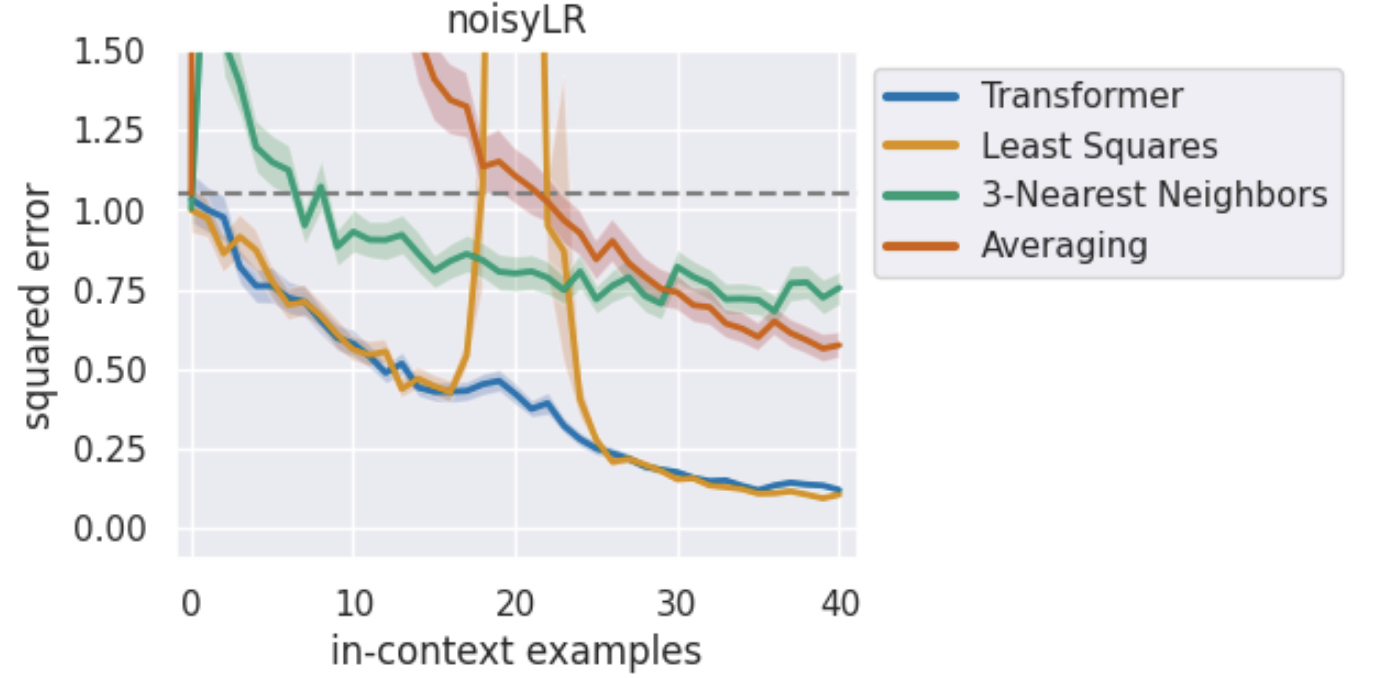}
    \caption{NoisyLR}
    \label{fig:sub18}
  \end{subfigure}
  \hfill
  \begin{subfigure}[t]{0.3\textwidth}
    \centering
    \includegraphics[width=\linewidth]{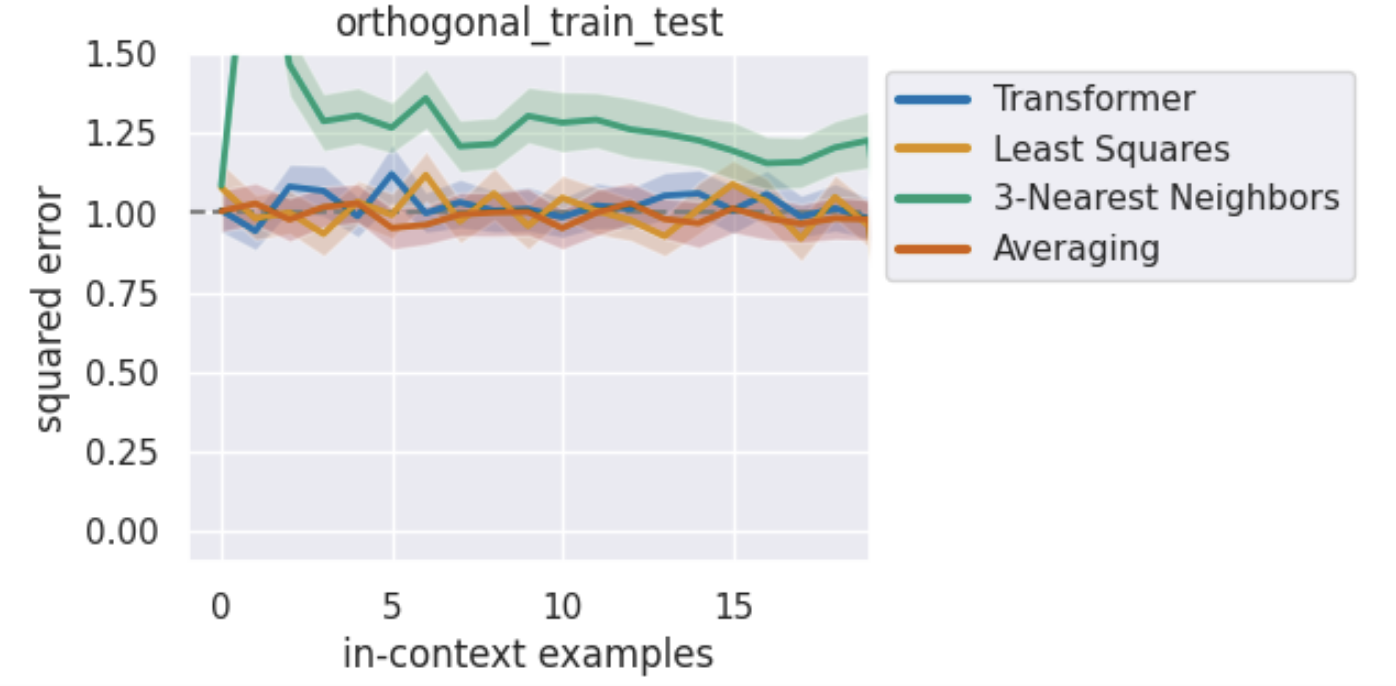}
    \caption{Orthogonal Sampling}
    \label{fig:sub19}
  \end{subfigure}

  \vspace{0.5cm}  

  \begin{subfigure}[t]{0.3\textwidth}
    \centering
    \includegraphics[width=\linewidth]{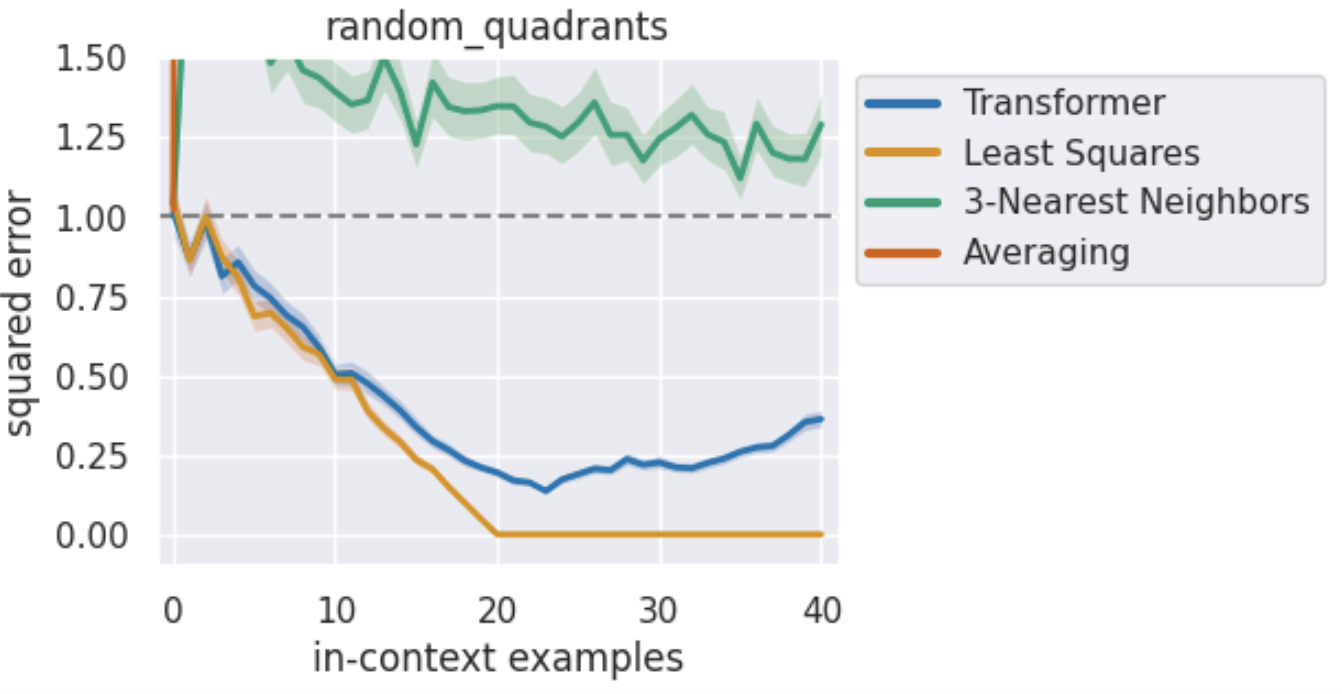}
    \caption{Random Quadrants}
    \label{fig:sub20}
  \end{subfigure}
  \hfill
  \begin{subfigure}[t]{0.3\textwidth}
    \centering
    \includegraphics[width=\linewidth]{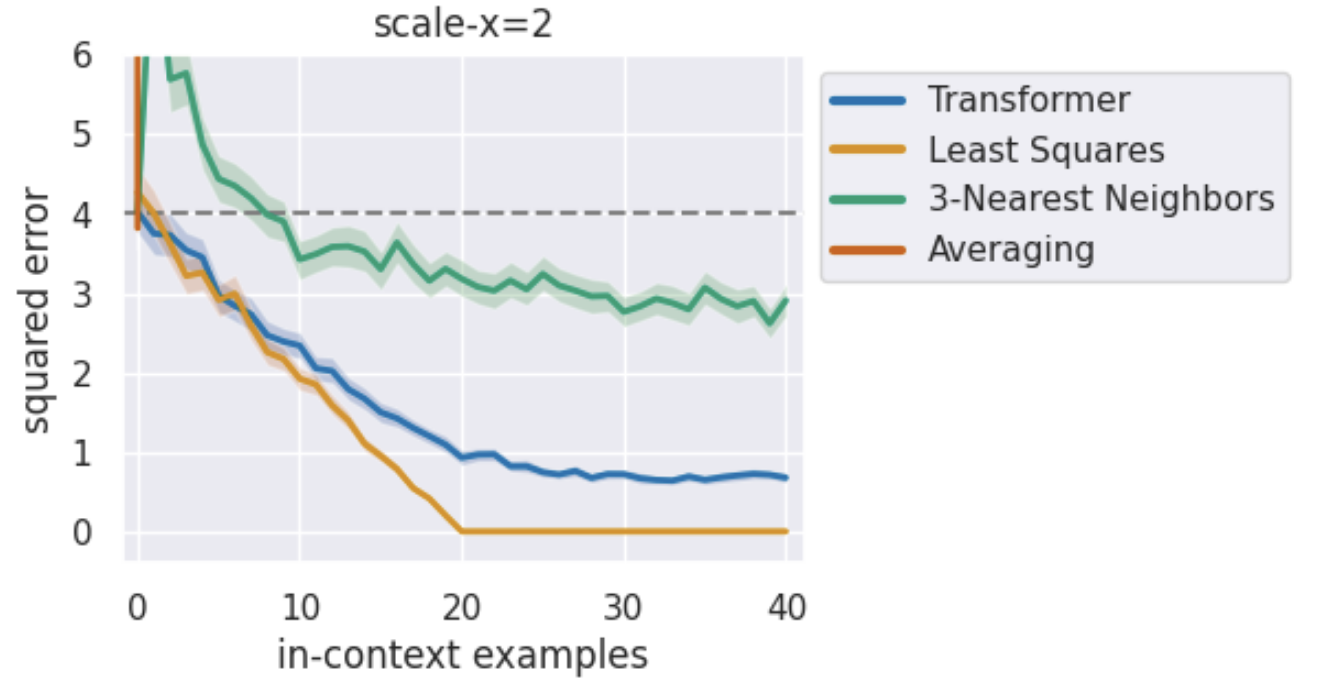}
    \caption{Scaled Distribution}
    \label{fig:sub21}
  \end{subfigure}
  \hfill
  \begin{subfigure}[t]{0.3\textwidth}
    \centering
    \includegraphics[width=\linewidth]{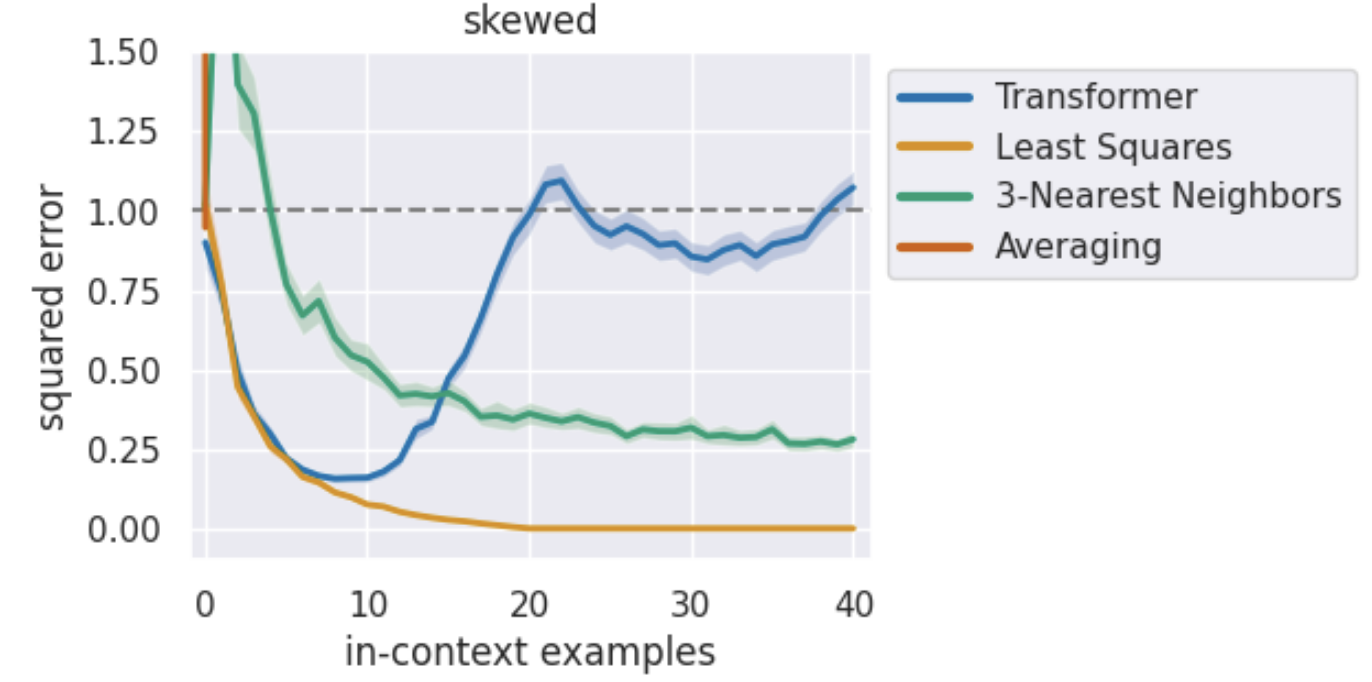}
    \caption{Skewed Distribution}
    \label{fig:sub22}
  \end{subfigure}

  \caption{Results of OOD Sampling on GPT2 with Flash Attention}
  \label{fig:six_results_2}
\end{figure}

\subsection{Hyena}
\begin{figure}[htbp]
  \centering

  \begin{subfigure}[t]{0.3\textwidth}
    \centering
    \includegraphics[width=\linewidth]{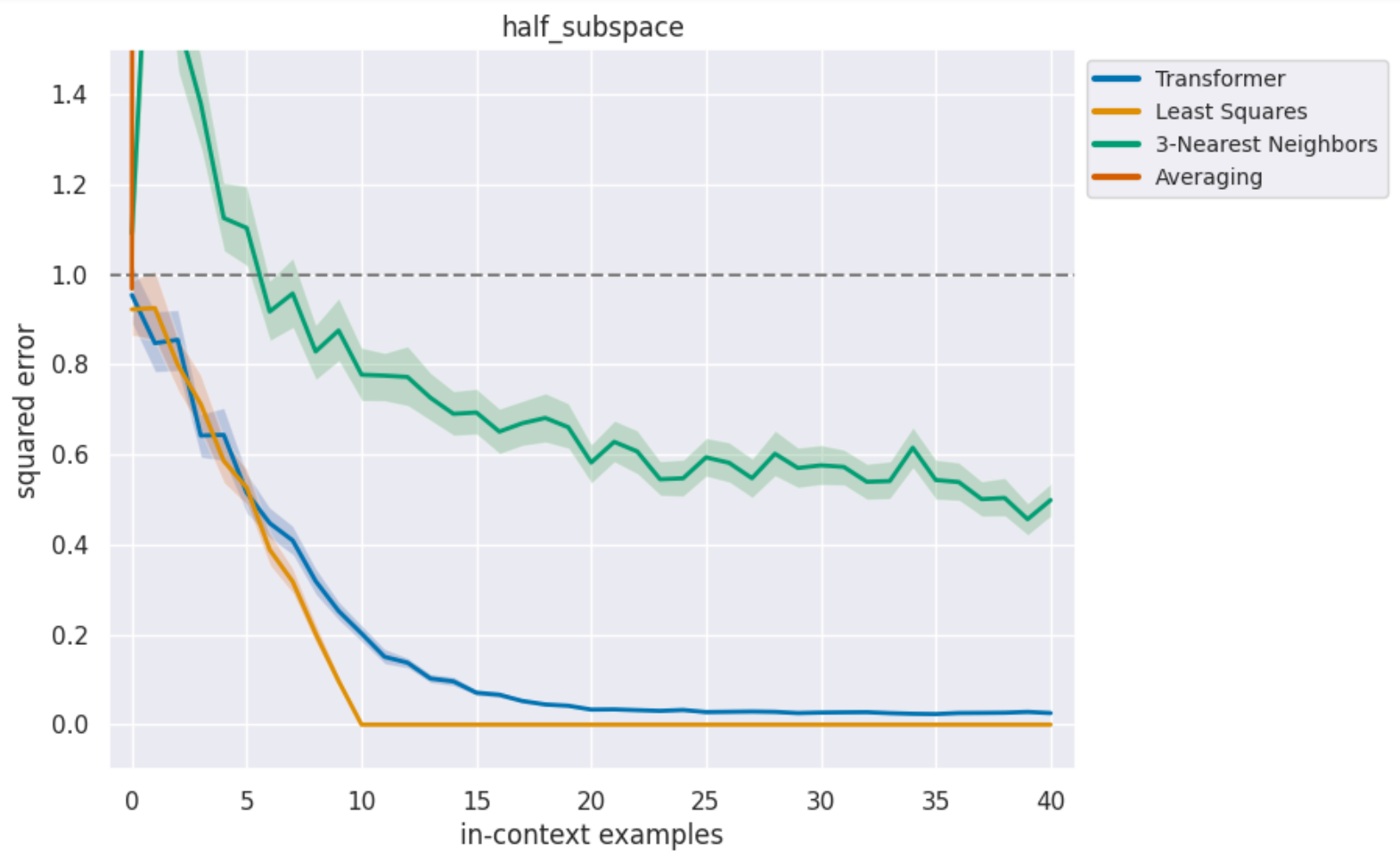}
    \caption{Half-subspace}
    \label{fig:sub23}
  \end{subfigure}
  \hfill
  \begin{subfigure}[t]{0.3\textwidth}
    \centering
    \includegraphics[width=\linewidth]{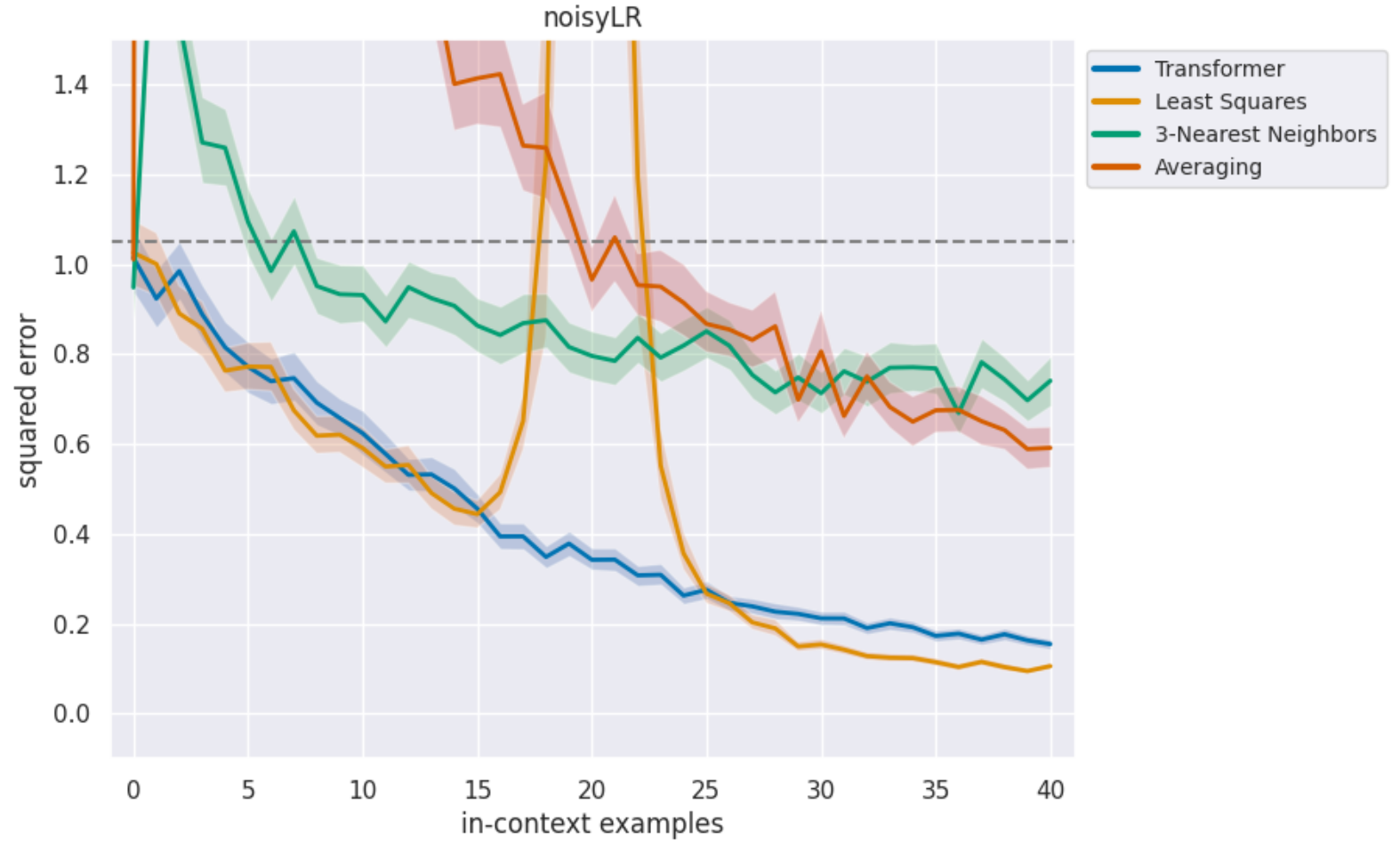}
    \caption{NoisyLR}
    \label{fig:sub24}
  \end{subfigure}
  \hfill
  \begin{subfigure}[t]{0.3\textwidth}
    \centering
    \includegraphics[width=\linewidth]{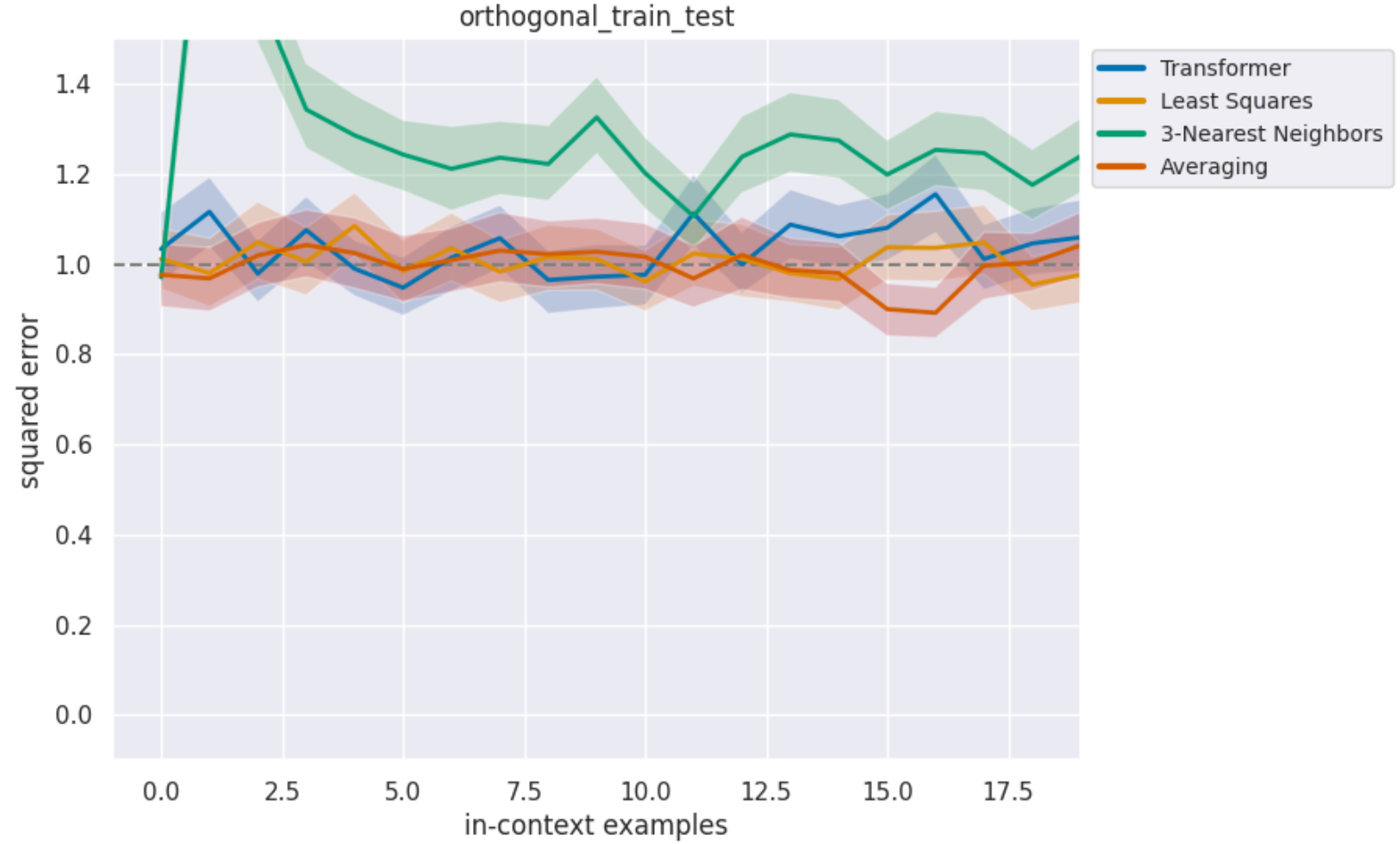}
    \caption{Orthogonal Sampling}
    \label{fig:sub25}
  \end{subfigure}

  \vspace{0.5cm}  

  \begin{subfigure}[t]{0.3\textwidth}
    \centering
    \includegraphics[width=\linewidth]{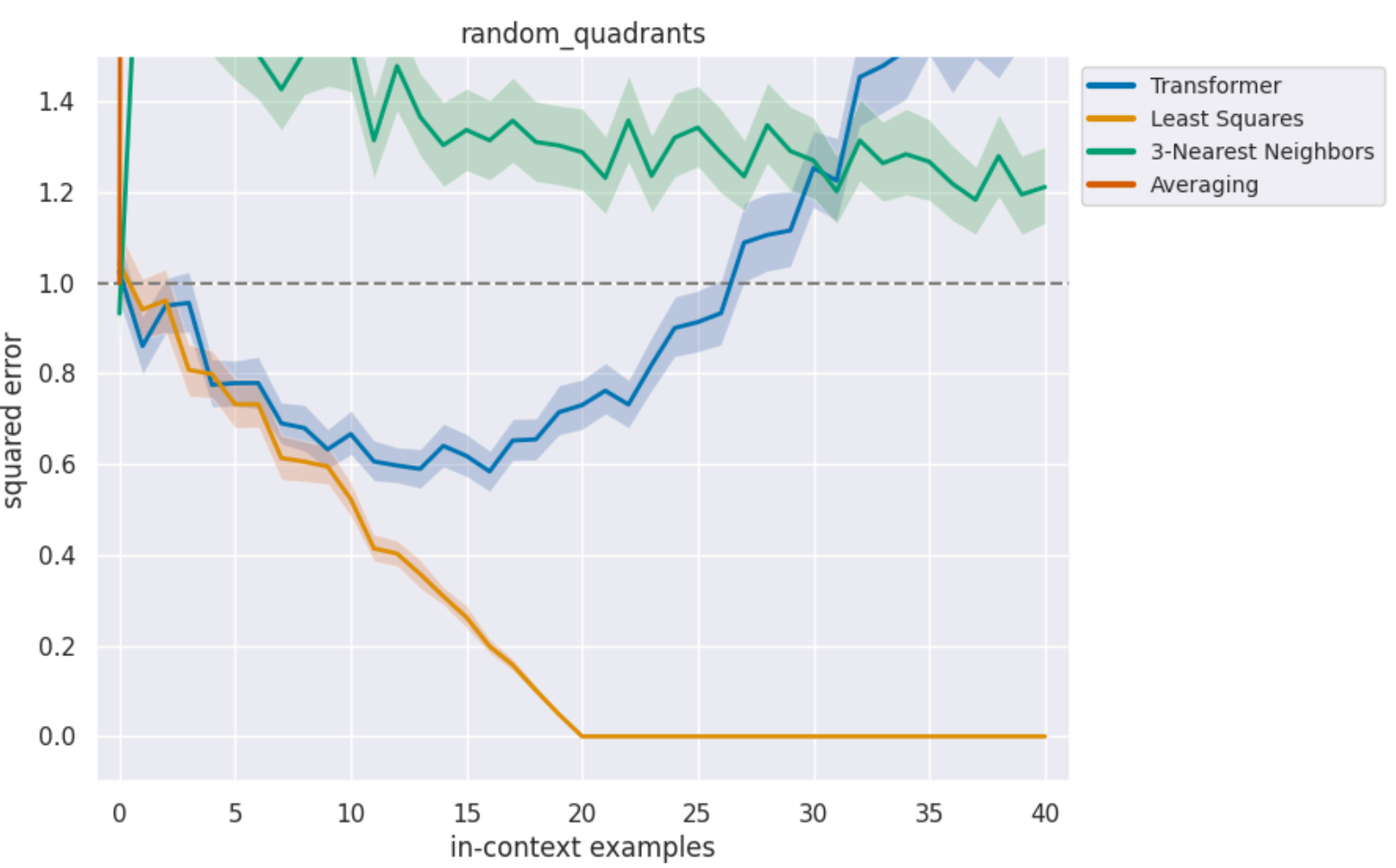}
    \caption{Random Quadrants}
    \label{fig:sub26}
  \end{subfigure}
  \hfill
  \begin{subfigure}[t]{0.3\textwidth}
    \centering
    \includegraphics[width=\linewidth]{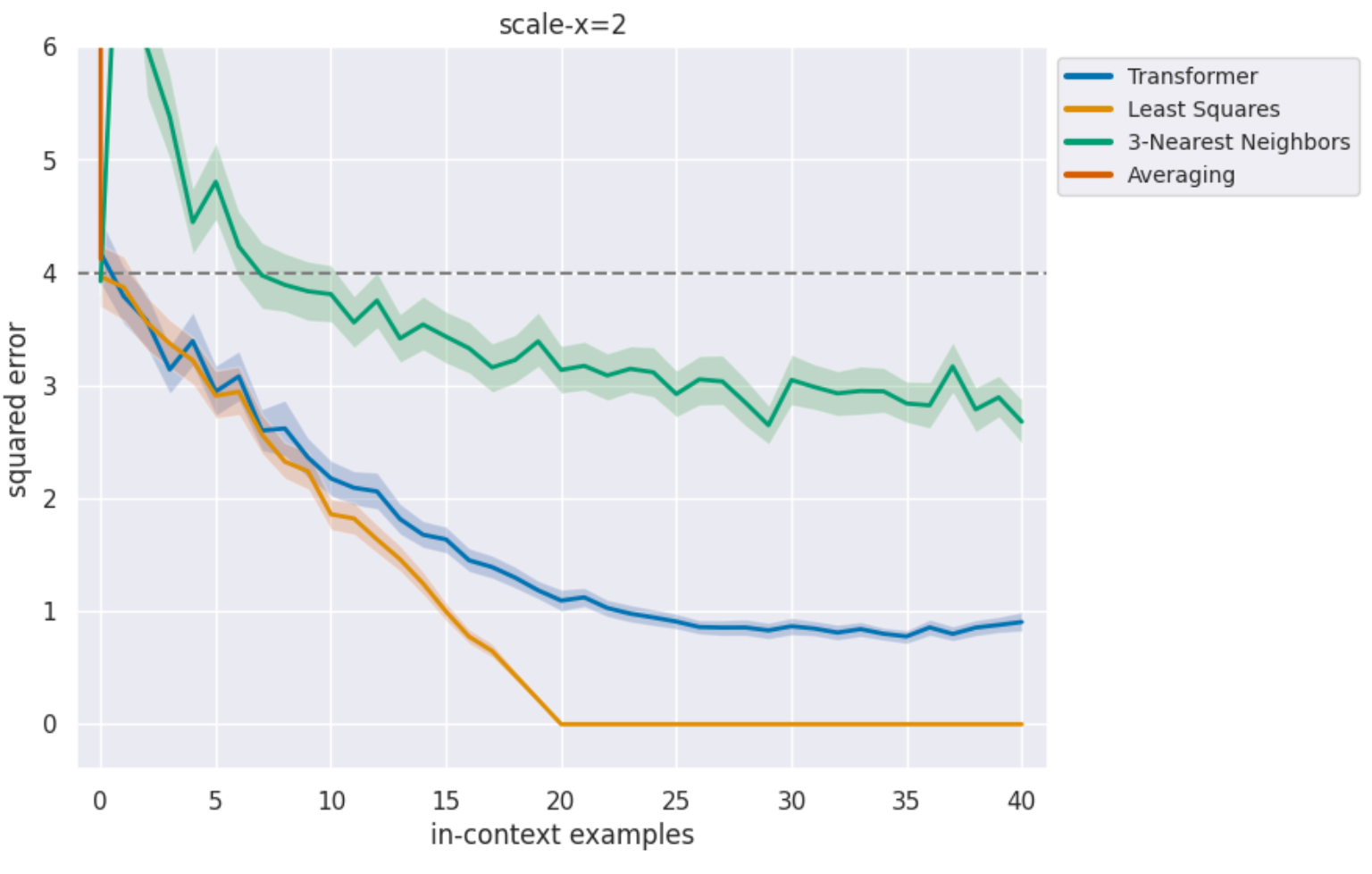}
    \caption{Scaled Distribution}
    \label{fig:sub27}
  \end{subfigure}
  \hfill
  \begin{subfigure}[t]{0.3\textwidth}
    \centering
    \includegraphics[width=\linewidth]{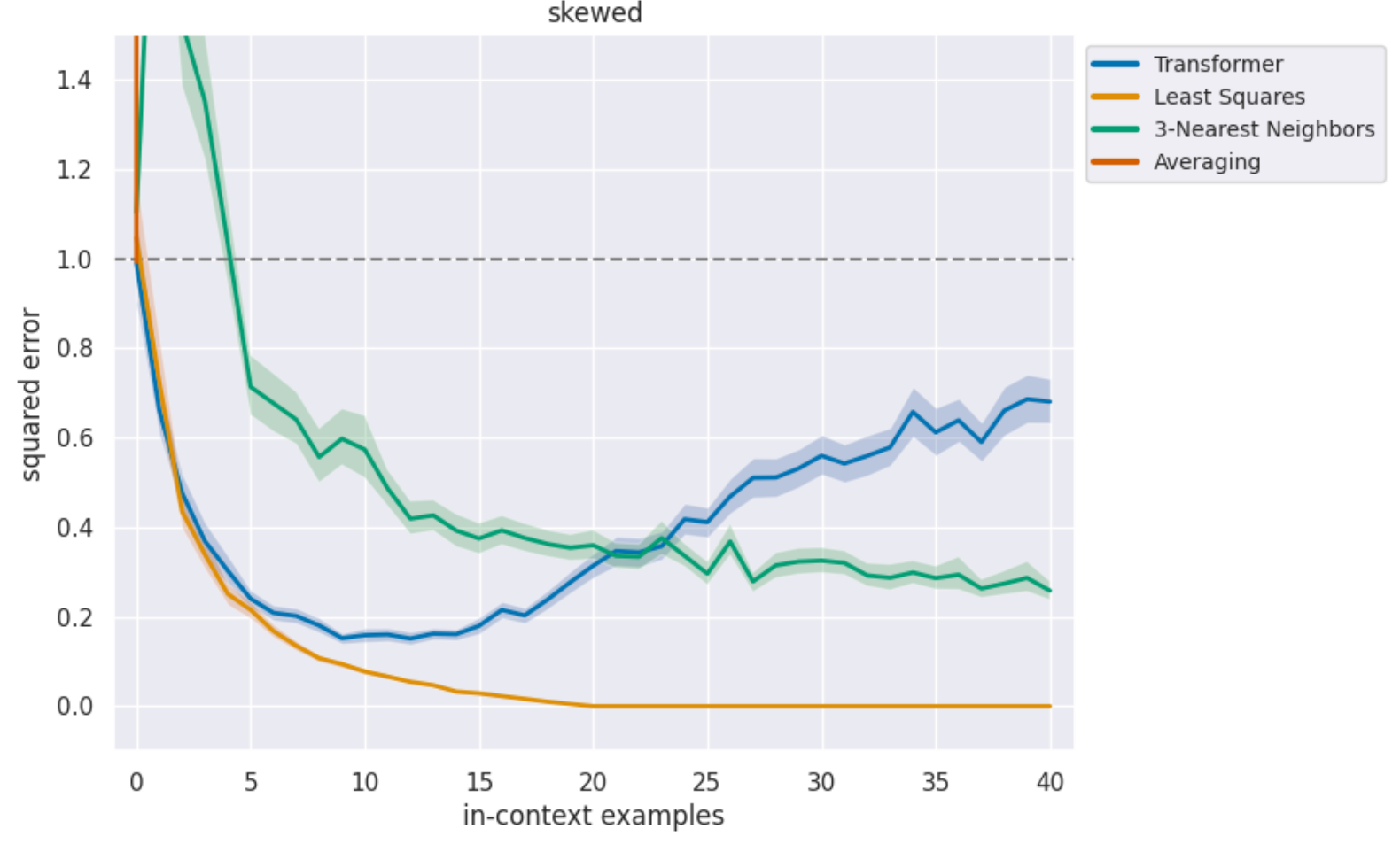}
    \caption{Skewed Distribution}
    \label{fig:sub28}
  \end{subfigure}

  \caption{Results of OOD Sampling on Hyena}
  \label{fig:six_results_2}
\end{figure}

\newpage
\subsection{Mamba}

\begin{figure}[htbp]
  \centering

  \begin{subfigure}[t]{0.3\textwidth}
    \centering
    \includegraphics[width=\linewidth]{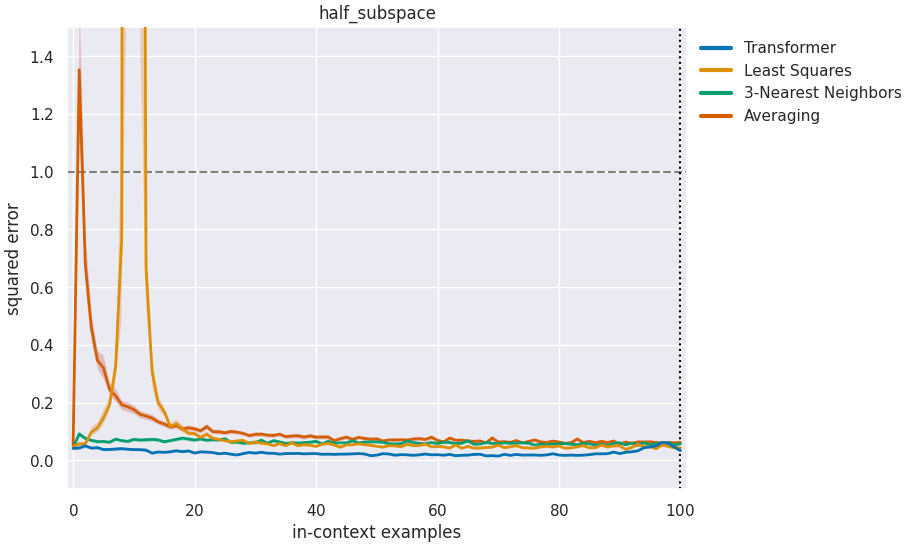}
    \caption{Half-subspace}
    \label{fig:sub29}
  \end{subfigure}
  \hfill
  \begin{subfigure}[t]{0.3\textwidth}
    \centering
    \includegraphics[width=\linewidth]{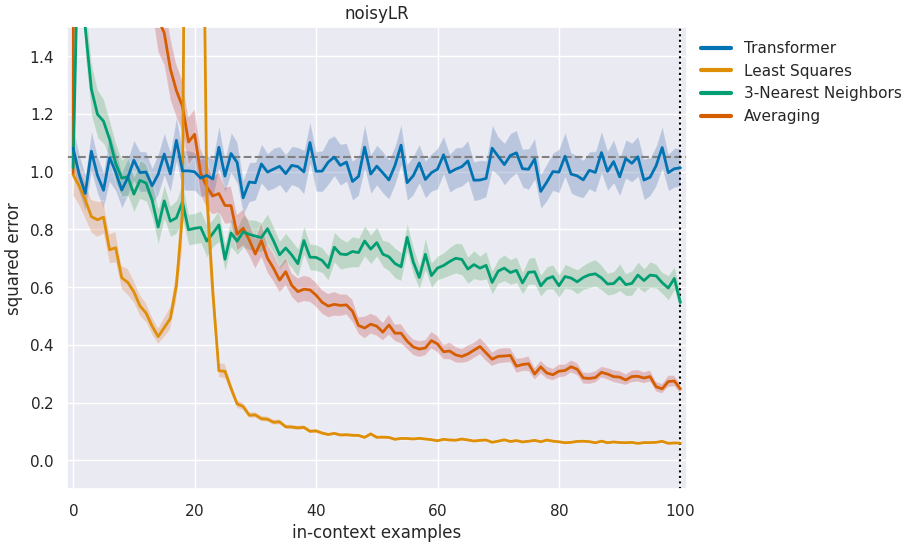}
    \caption{NoisyLR}
    \label{fig:sub30}
  \end{subfigure}
  \hfill
  \begin{subfigure}[t]{0.3\textwidth}
    \centering
    \includegraphics[width=\linewidth]{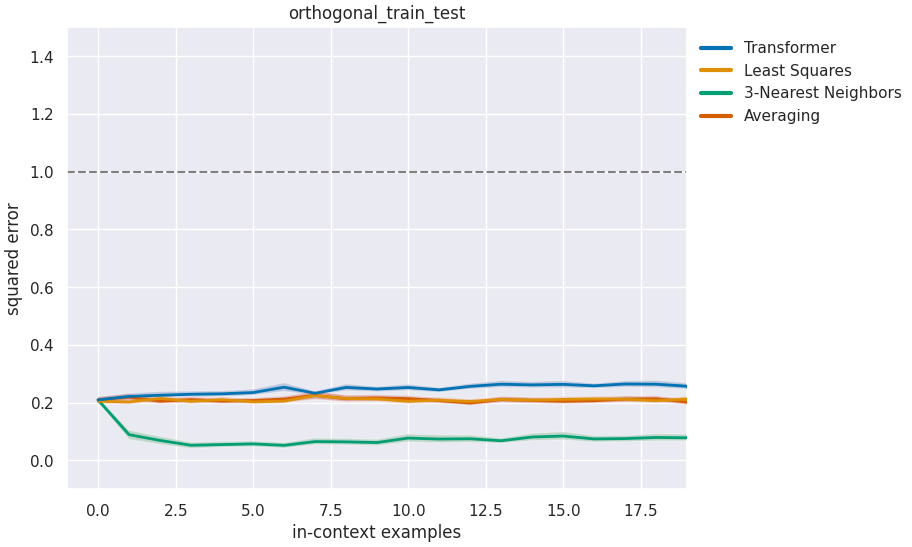}
    \caption{Orthogonal Sampling}
    \label{fig:sub31}
  \end{subfigure}

  \vspace{0.5cm}  

  \begin{subfigure}[t]{0.3\textwidth}
    \centering
    \includegraphics[width=\linewidth]{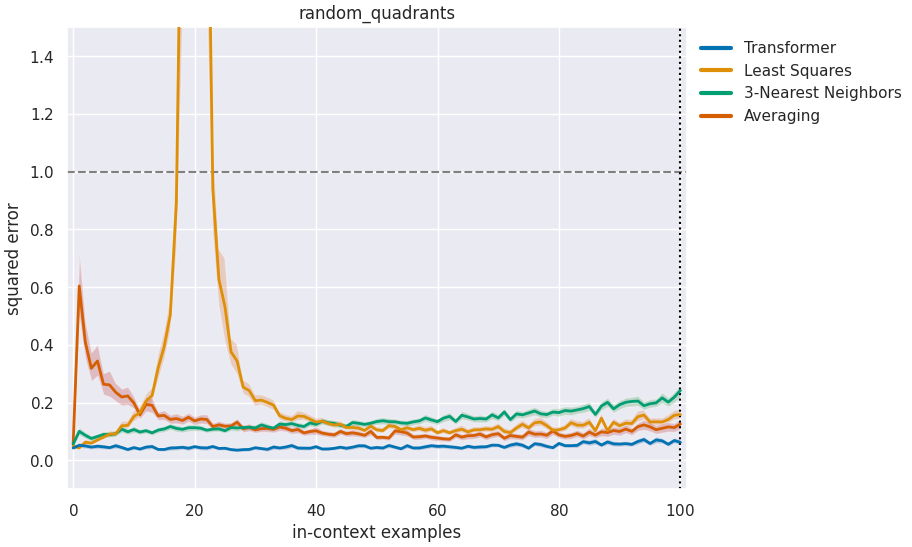}
    \caption{Random Quadrants}
    \label{fig:sub32}
  \end{subfigure}
  \hfill
  \begin{subfigure}[t]{0.3\textwidth}
    \centering
    \includegraphics[width=\linewidth]{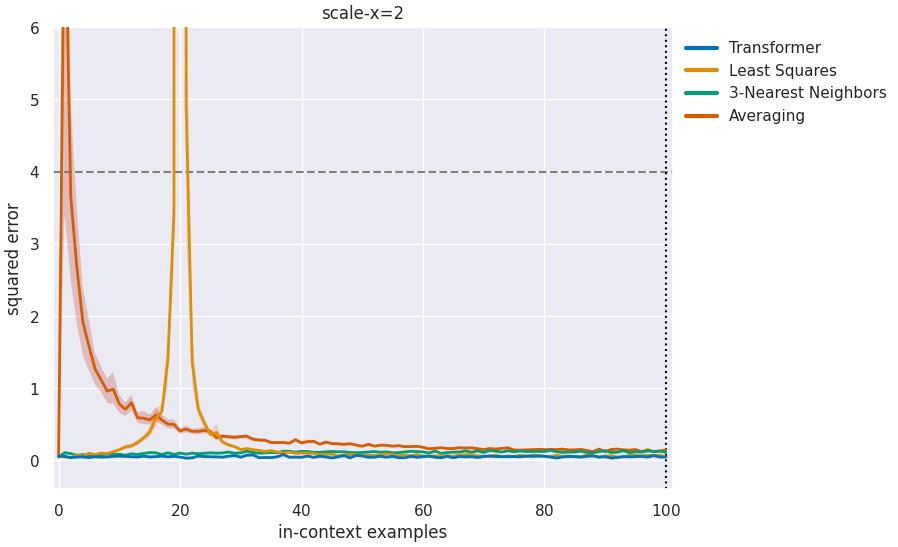}
    \caption{Scaled Distribution}
    \label{fig:sub33}
  \end{subfigure}
  \hfill
  \begin{subfigure}[t]{0.3\textwidth}
    \centering
    \includegraphics[width=\linewidth]{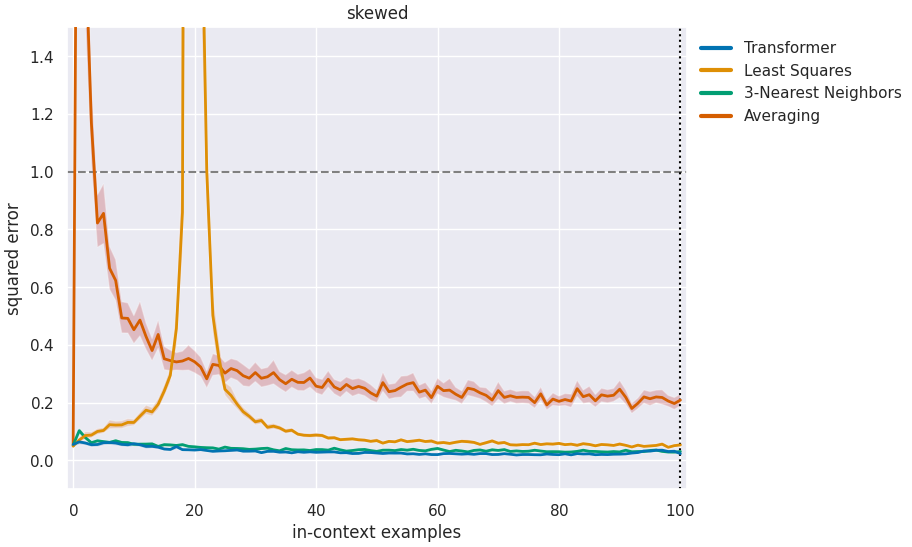}
    \caption{Skewed Distribution}
    \label{fig:sub34}
  \end{subfigure}

  \caption{Results of OOD Sampling on Mamba (Gaussian Kernel Regression)}
  \label{fig:six_results_3}
\end{figure}

\begin{figure}[htbp]
  \centering

  \begin{subfigure}[t]{0.3\textwidth}
    \centering
    \includegraphics[width=\linewidth]{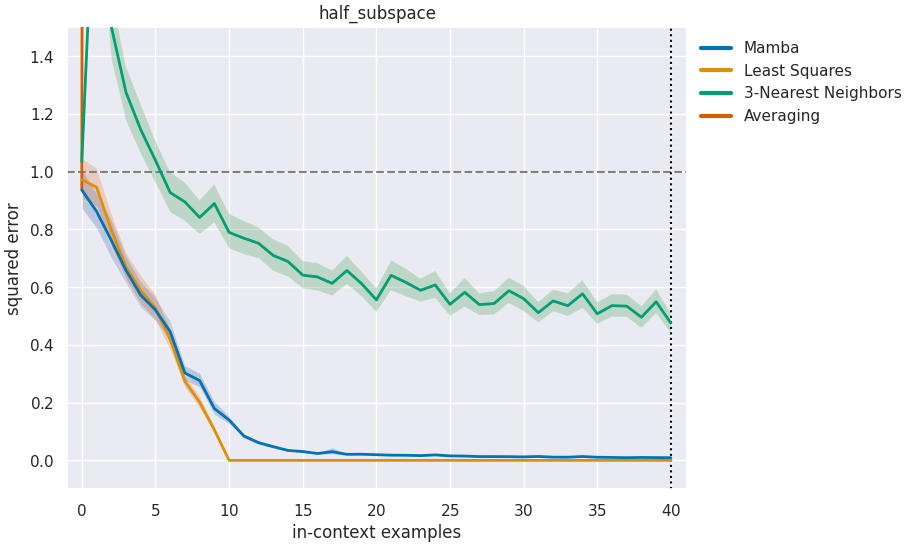}
    \caption{Half-subspace}
    \label{fig:sub35}
  \end{subfigure}
  \hfill
  \begin{subfigure}[t]{0.3\textwidth}
    \centering
    \includegraphics[width=\linewidth]{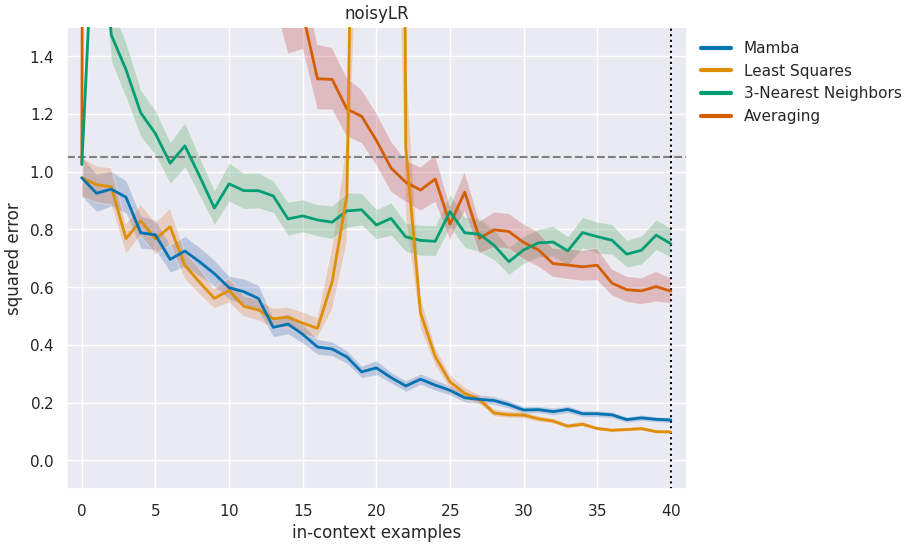}
    \caption{NoisyLR}
    \label{fig:sub36}
  \end{subfigure}
  \hfill
  \begin{subfigure}[t]{0.3\textwidth}
    \centering
    \includegraphics[width=\linewidth]{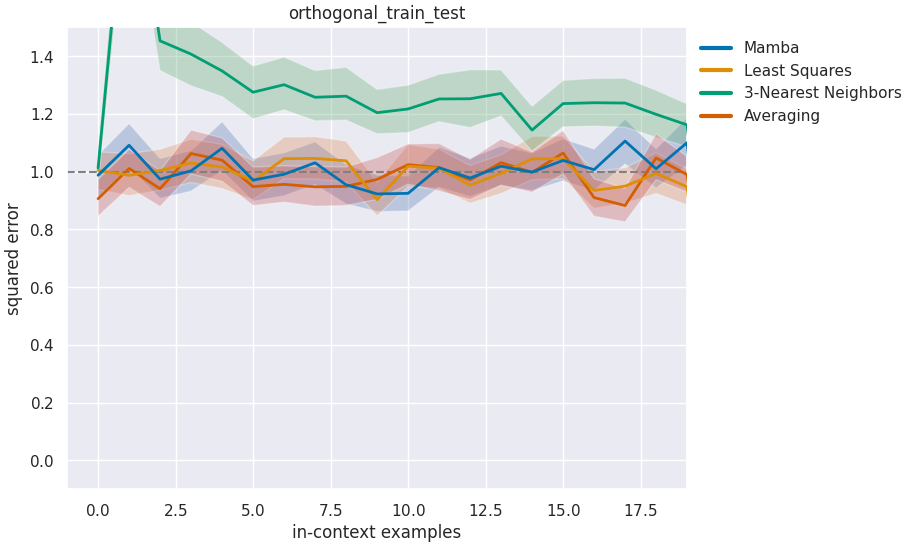}
    \caption{Orthogonal Sampling}
    \label{fig:sub37}
  \end{subfigure}

  \vspace{0.5cm}  

  \begin{subfigure}[t]{0.3\textwidth}
    \centering
    \includegraphics[width=\linewidth]{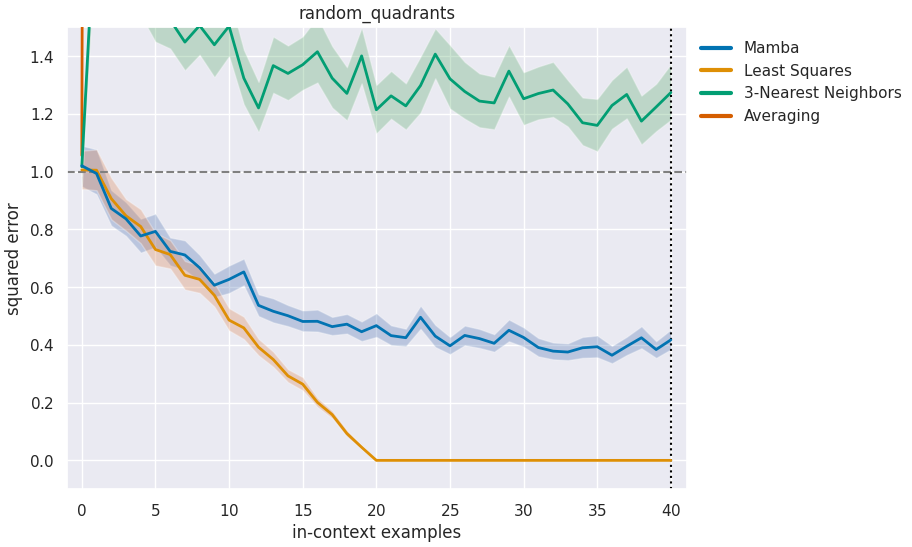}
    \caption{Random Quadrants}
    \label{fig:sub38}
  \end{subfigure}
  \hfill
  \begin{subfigure}[t]{0.3\textwidth}
    \centering
    \includegraphics[width=\linewidth]{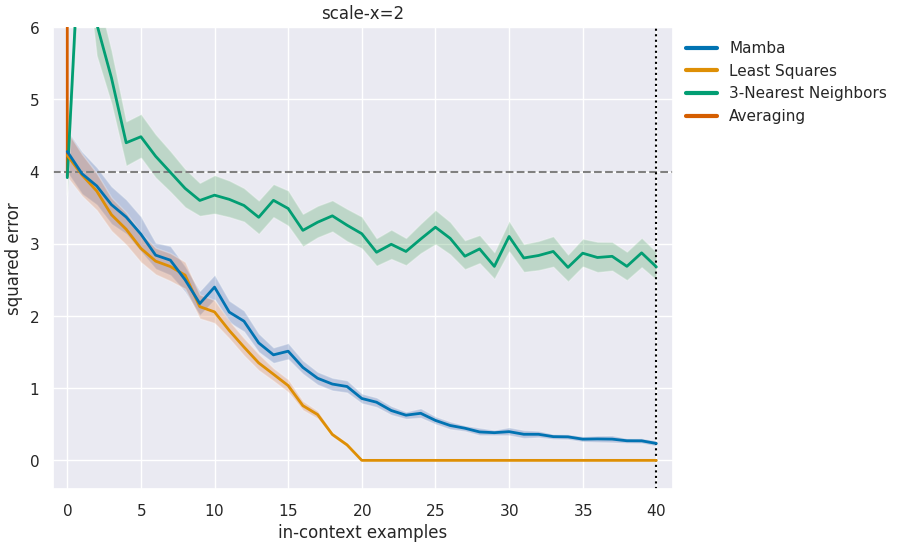}
    \caption{Scaled Distribution}
    \label{fig:sub39}
  \end{subfigure}
  \hfill
  \begin{subfigure}[t]{0.3\textwidth}
    \centering
    \includegraphics[width=\linewidth]{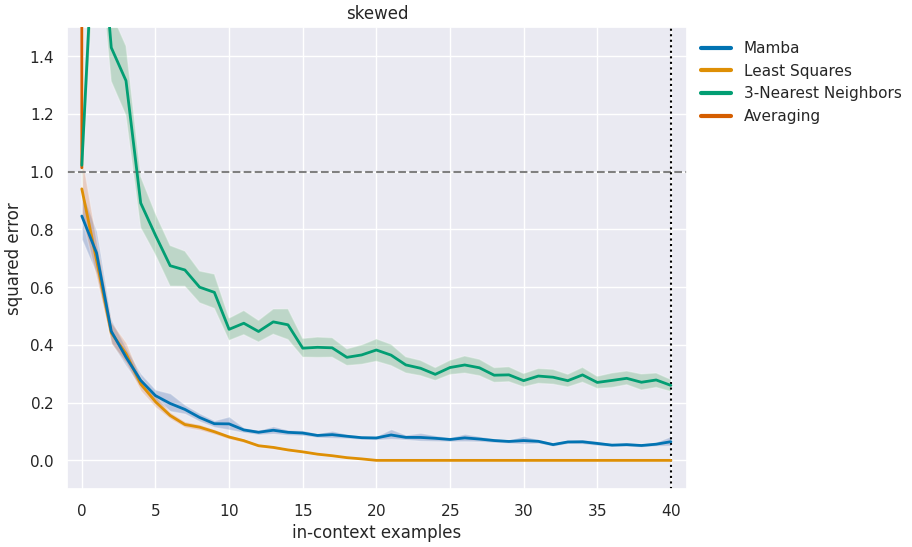}
    \caption{Skewed Distribution}
    \label{fig:sub40}
  \end{subfigure}

  \caption{Results of OOD Sampling on Mamba (Nonlinear Dynamics)}
  \label{fig:six_results_2}
\end{figure}



\newpage
\section{Additional Graphs for Discussion}

\begin{figure}[h]
    \centering
    \includegraphics[width=0.7\linewidth]{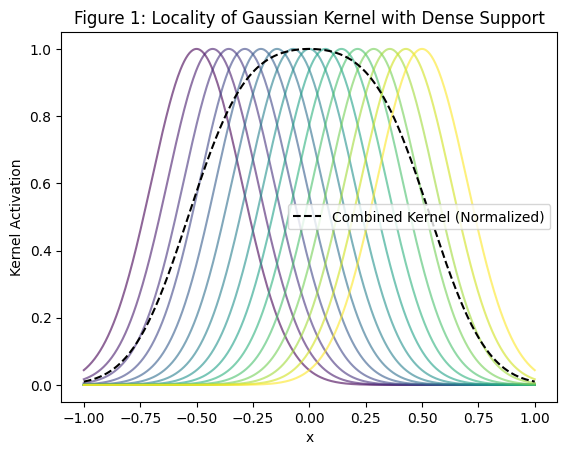}
    \caption{Support and query points in $[-1, 1]$ under Gaussian kernel similarity. Overlapping response regions cause trivial local interpolation.}
    \label{fig:kernel_locality}
\end{figure}

\begin{figure}[h]
    \centering
    \includegraphics[width=0.7\linewidth]{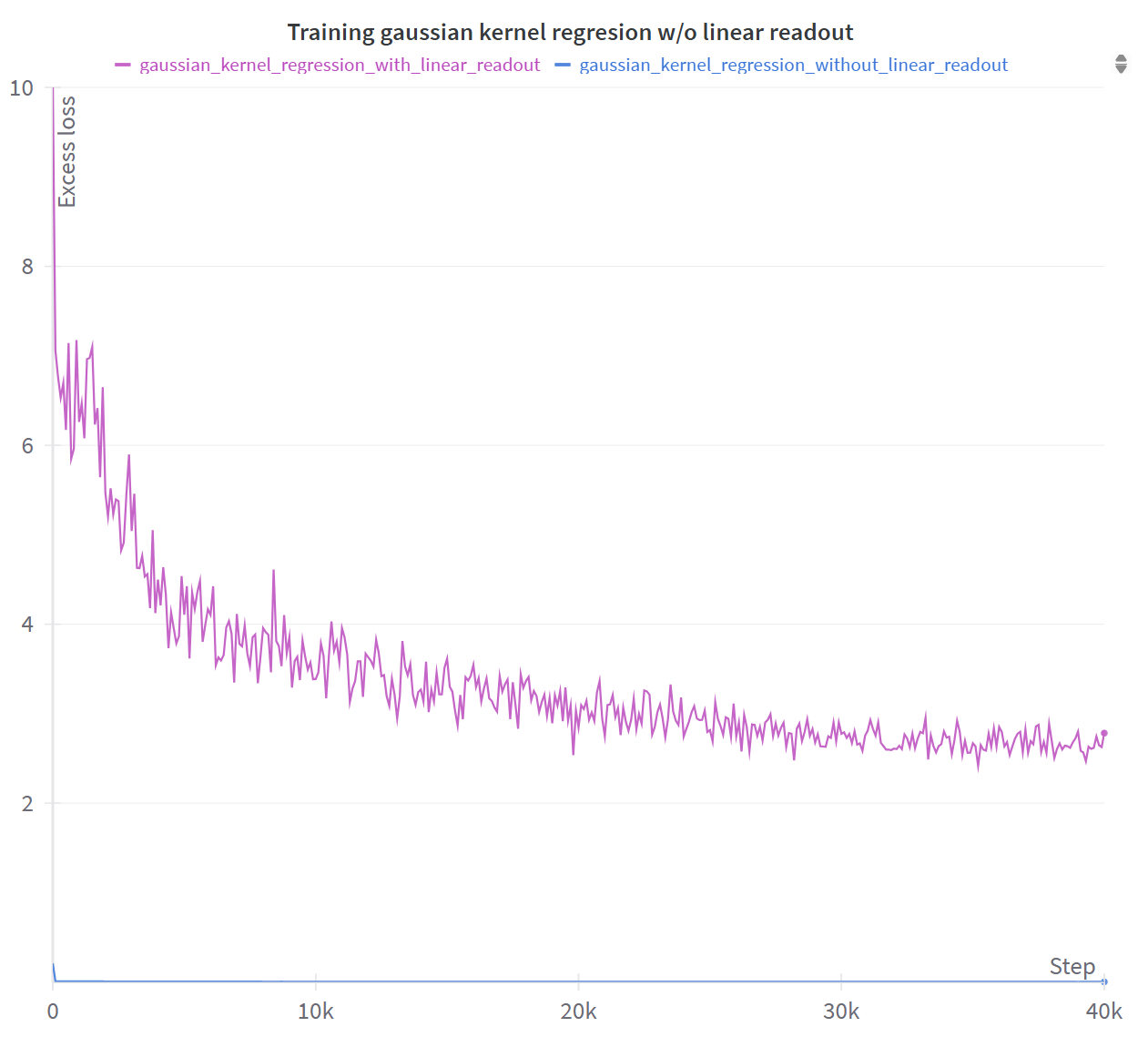}
    \caption{Training curves with/without linear readout. Naive kernel variant shows flat loss; readout recovers learning signal.}
    \label{fig:readout_loss}
\end{figure}

\begin{figure}[h]
    \centering
    \includegraphics[width=0.7\linewidth]{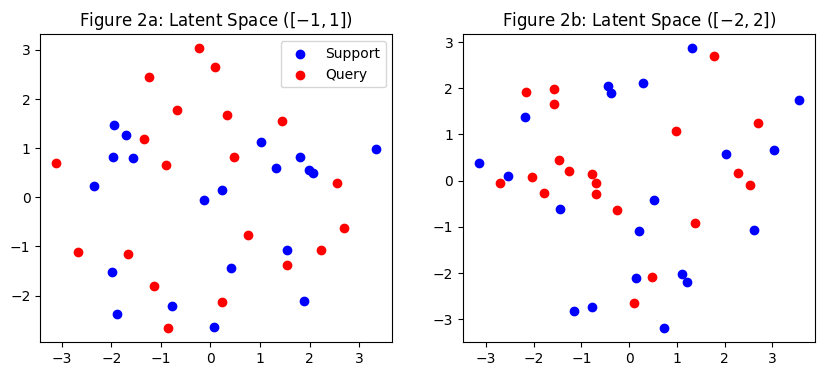}
    \caption{Latent representation separation (t-SNE) under different input ranges. Scaling to $[-2,2]$ improves support-query contrast.}
    \label{fig:tsne_latents}
\end{figure}

\begin{figure}[h]
    \centering
    \includegraphics[width=0.7\linewidth]{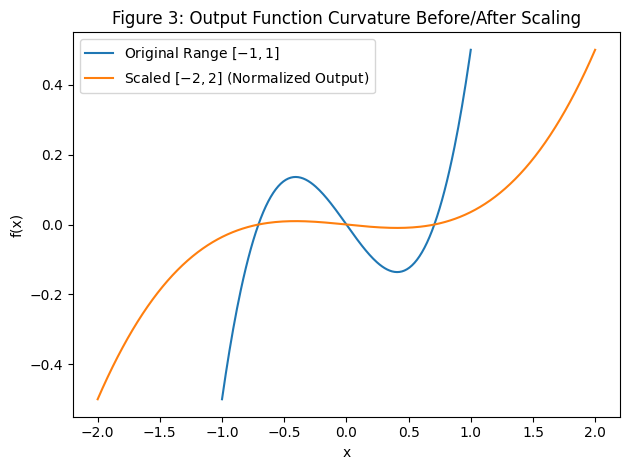}
    \caption{Function curves before/after input scaling. Larger input domains enhance curvature, aiding pattern recognition.}
    \label{fig:fx_scaling}
\end{figure}

\begin{figure}[h]
    \centering
    \includegraphics[width=0.7\linewidth]{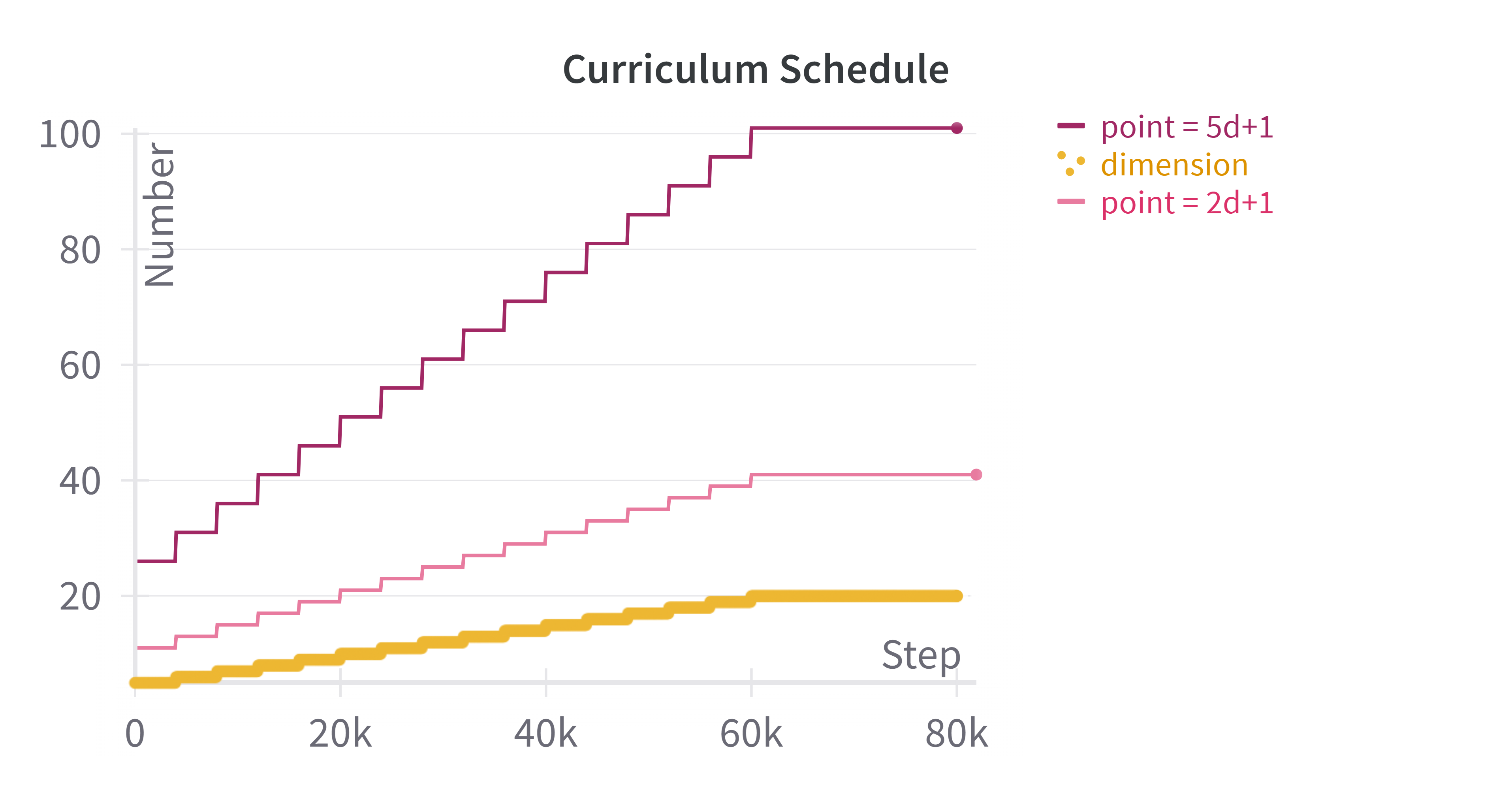}
    \caption{Curriculum schedule by task type. Complex tasks (NN, trees) require more context tokens per dimension.}
    \label{fig:curriculum_schedule}
\end{figure}

\end{document}